%% file: main.tex
\pdfoutput=1

\documentclass[a4paper, 10pt, conference]{ieeeconf}      

\IEEEoverridecommandlockouts                              
\title{\LARGE \bf
Skeleton Driven Non-rigid Motion Tracking and 3D Reconstruction }

\author{Shafeeq Elanattil$^{1,2}$, Peyman Moghadam$^{1,2}$, 
Simon Denman$^{2}$, Sridha Sridharan$^{2}$, Clinton Fookes$^{2}$
\thanks{$^{1}$ The authors are with the Robotics and Autonomous Systems, CSIRO, DATA61, Brisbane, QLD 4069, Australia.
E-mails: {\tt\small \emph{firstname.lastname}@data61.csiro.au}}%
\thanks{$^{2}$ The authors are with the School of Electrical Engineering and Computer
Science, Queensland University of Technology (QUT), Brisbane, Australia.
E-mails: {\tt\small \emph{\{shafeeq.elanattil, peyman.moghadam, s.denman, s.sridharan, c.fookes}\}@qut.edu.au}}%
}

 \PassOptionsToPackage{bookmarks=false}{hyperref}
\usepackage{times}
\usepackage{epsfig}
\usepackage{graphicx}
\usepackage{amsmath}
\usepackage{amssymb}
\usepackage{balance}
\usepackage{caption}

\usepackage[utf8]{inputenc}
\usepackage{dirtytalk}
\usepackage{booktabs}
\usepackage{subfig}
\usepackage{svg}
\usepackage{amsmath}
\usepackage{epstopdf}
\usepackage{graphicx,import}
\usepackage{gensymb}
\let\labelindent\relax
\usepackage{enumitem}
\usepackage{algorithm}
\usepackage[noend]{algpseudocode}
\usepackage{amsfonts}

\usepackage{amsthm}
\usepackage{bm} 
\usepackage{textcomp}
\usepackage[percent]{overpic}
\usepackage{dsfont}
\usepackage{mathtools}
\usepackage[%
  colorlinks=true,%
  urlcolor=red%
]{hyperref}%
\usepackage{multirow}
\usepackage{float}
\usepackage{subfig}
\usepackage{stackengine}

\graphicspath{{./figures/}}
\newcommand{\etal}{\emph{et~al.}}

\newcommand\myeq{\stackrel{\mathclap{\normalfont\mbox{def}}}{=}}

\emergencystretch=\maxdimen
\begin{document}

\maketitle
\thispagestyle{empty}
\pagestyle{empty}

\input{sections/abstract.tex}
\input{sections/introduction.tex}

\input{sections/literature_review.tex}
\input{sections/methodology.tex}

\input{sections/synthetic_data.tex}

\input{sections/experiments.tex}

\input{sections/conclusion.tex}




\balance
\bibliographystyle{IEEEtran}
\bibliography{publications}

\end{document}

%% file: sections/abstract.tex
\begin{abstract}

This paper presents a method which can track and 3D reconstruct the non-rigid surface motion of human performance using a moving RGB-D camera. 3D reconstruction of marker-less human performance is a challenging problem due to the large range of articulated motions and considerable non-rigid deformations. Current approaches use local optimization for tracking. These methods need many iterations to converge and may get stuck in local minima during sudden articulated movements. We propose a puppet model-based tracking approach using skeleton prior, which provides a better initialization for tracking articulated movements. The proposed approach uses an aligned puppet model to estimate correct correspondences for human performance capture. We also contribute a synthetic dataset which provides ground truth locations for frame-by-frame geometry and skeleton joints of human subjects. Experimental results show that our approach is more robust when faced with sudden articulated motions, and provides better 3D reconstruction compared to the existing state-of-the-art approaches.

\end{abstract}

%% file: sections/introduction.tex
\section{Introduction}

The goal of marker-less performance capture is to track the motion of a moving human and reconstruct a temporally coherent representation of its dynamically deforming surfaces. The large range of motions and considerable non-rigid deformations of the human body make this problem challenging, even when multiple views are available. The reconstruction from a single-view is further complicated because of self-occlusions. Despite all of these challenges, a solution to this problem is a necessity for a broad range of applications including computer animation, visual effects, and free-viewpoint video, medicine and biomechanics.

\begin{figure}[t]
\begin{center}
   \includegraphics[width=\linewidth, height =7cm]{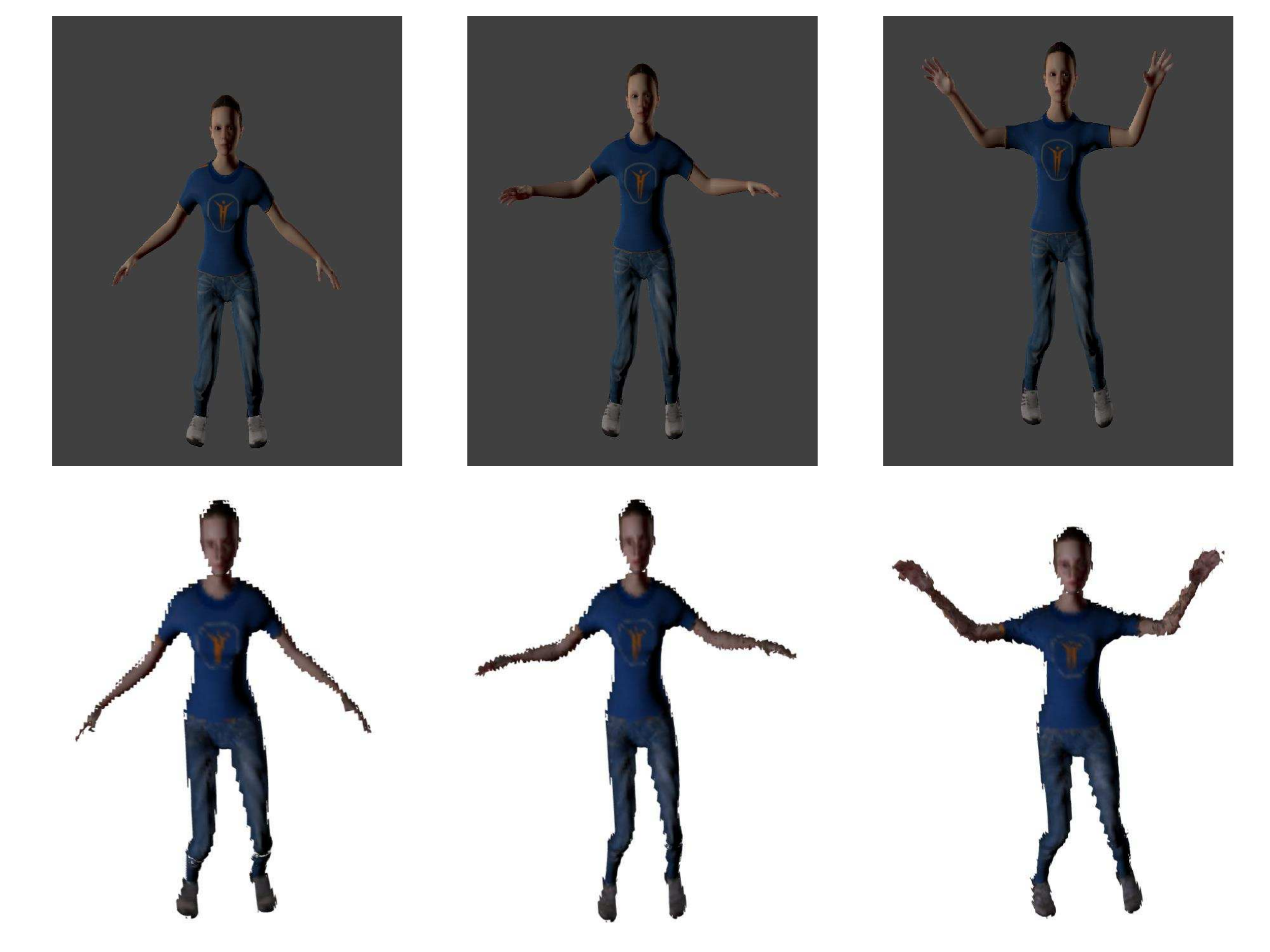}
\end{center}
 \caption{Qualitative results of live reconstruction from \textquoteleft Exercise\textquoteright \hspace{1pt} data sequence in our synthetic dataset. The upper row corresponds to images of different frame indices and the lower row shows respective 3D reconstructions.}
 \label{QualitativeExerciseRecon}
\end{figure}

Recently, volumetric depth fusion methods \cite{newcombe2015dynamicfusion,innmann2016volumedeform} have enabled reconstruction systems to bypass the need for a complete model of the object by incrementally reconstructing model. However, these approaches do not consider the intrinsic articulated nature of the human body, and thus fail to reconstruct human subjects when they undergo natural articulated movements. Yu \etal{}~\cite{yu122017bodyfusion} extended this approach to reconstruct the non-rigid surface motion of human performance by using an underlying skeleton prior. 
These volumetric depth fusion methods \cite{newcombe2015dynamicfusion,innmann2016volumedeform,yu122017bodyfusion, vidas20133d, vidassenors} track the model from incoming data using local optimization and are thus prone to becoming trapped in local minima. Moreover, they use projective correspondences for tracking which often gives erroneous correspondences during natural human motions. Therefore these approaches often fail when tracking sudden articulated motion sequences.

In this paper, we propose a method which uses skeleton prior in each frame to robustly reconstruct human performance. Our approach uses a human puppet model for motion tracking based on the detected skeleton. For each frame, the puppet model is aligned with incoming depth data by estimating the rigid transformation of each body part. The rigid transformation parameters provide a perfect initialization for tracking and help avoid the optimization becoming stuck in local minima. In addition, the aligned puppet model is used to estimate correct correspondences for tracking. Figure \ref{QualitativeExerciseRecon} shows an example of our 3D reconstruction results at different frame instances. We have also developed a synthetic dataset for evaluating RGB-D based methods for reconstructing humans under articulated motion. The key contributions of our approach can be summarized as follows:   

\begin{itemize}

\item We proposed a puppet model-based tracking approach using skeleton prior and show how this can be used to provide a better initialization for tracking articulated movements. 

\item Our approach shows how puppet model-based tracking is used to estimate correct correspondences for 3D reconstruction of human performance.  
\item We propose a synthetic dataset which provides ground truth for frame-by-frame geometry and skeleton joints of human subjects. This enables per-frame quantitative evaluation for 3D reconstruction of non-rigid human movements.

\end{itemize}

%% file: sections/literature_review.tex
\section{Related Works}

Non-rigid 3D reconstruction approaches from RGB-D data can be classified into two categories. The first category uses priors as a template or multiple cameras to make the problem tractable. The second category incrementally reconstructs the scene without any template (template-free). In this section, we briefly summaries recent related works. 

\subsection{Multi-view and Template based approaches}

Vlasic \etal{}~\cite{vlasic2009dynamic} uses a skeleton based human model for tracking articulated motion in a multi-view setting. Even though the parameter space is reduced to the joint angles, this limits the range of non-rigid deformation that can be modeled. Zollh{\"o}fer \etal{}~\cite{zollhofer2014real} proposed a deformation graph based non-rigid tracking method using a model. They used a non-rigid Iterative Closest Point (ICP) approach for tracking. Guo \etal{}~\cite{l0Norigid2015} presented a motion tracking approach by using L0 optimization to robustly reconstruct non-rigid geometry using a single depth sensor. Although these techniques achieve accurate non-rigid tracking for a wide variety of motions, they require an initial template geometry as a prior. For this purpose, the non-rigid object has to be still during template generation, which cannot be guaranteed in general situations. 

Another class of methods \cite{loper2015smpl} \cite{paulus2015augmented} need to first learn a parametric model of the target object and use this model for fitting the data. However, these methods fail to reconstruct objects which cannot be represented by the dataset. Bogo \etal{}~\cite{bogo2014faust} used a parametric human model to track a moving person. Because the parametric model is developed from a large 3D dataset of undressed human bodies, their method could not reconstruct dressed human bodies. Recently Duo \etal{}~\cite{dou2016fusion4d} demonstrated a non-rigid reconstruction system using 24 cameras and multiple GPUs, which is a setup not available to a general user.

\subsection{Template-free approaches}

Template-free approaches incrementally reconstruct the object by tracking motion simultaneously. This kind of setup is desirable for hand-held scanning based systems. Dou \etal{}~\cite{dou20153d} used a non-rigid bundle adjustment for reconstructing a non-rigid object; however, the method takes 9 to 10 hours for optimization. DynamicFusion was the first approach to simultaneously reconstruct and track a non-rigid scene in real-time \cite{newcombe2015dynamicfusion}. VolumeDeform \cite{innmann2016volumedeform} extended this work by using SIFT features across all images to reduce the drift. Both of the approaches provide compelling results for relatively slow motions. 

Guo \etal{}~\cite{guo2017real} improved tracking by using surface albedo, which is estimated using lighting coefficients under a Lambertian surface assumption. However, the Lambertian surface assumption only works in constrained lighting conditions. Slacheva \etal{}~\cite{slavcheva2017killingfusion} proposed a Signed Distance Function (SDF) based flow vector field for non rigid tracking. This approach can track a greater range of motions and solve problems due to topological changes up-to some extent. However, SDF based optimization significantly increases the computational overhead. Therefore this approach can only produce a coarse reconstruction in real-time. 

Elanattil \etal{}~\cite{elanattil2018non} proposed a method that uses camera pose estimated from the background to improve robustness in handling larger frame-to-frame motions. They also use a multi-scale deformation architecture which enables a wider range of tracking. However, the projective correspondence based tracking fails during articulated motions. BodyFusion~\cite{yu122017bodyfusion} is the more closely related to our work. They incrementally reconstruct the human subject using a skeleton prior during optimization for tracking. The key difference in our approach is that we are using a human model for tracking by using skeleton joint positions at each frame. This helps our approach track and reconstruct human subjects during sudden articulated movements. 

%% file: sections/methodology.tex
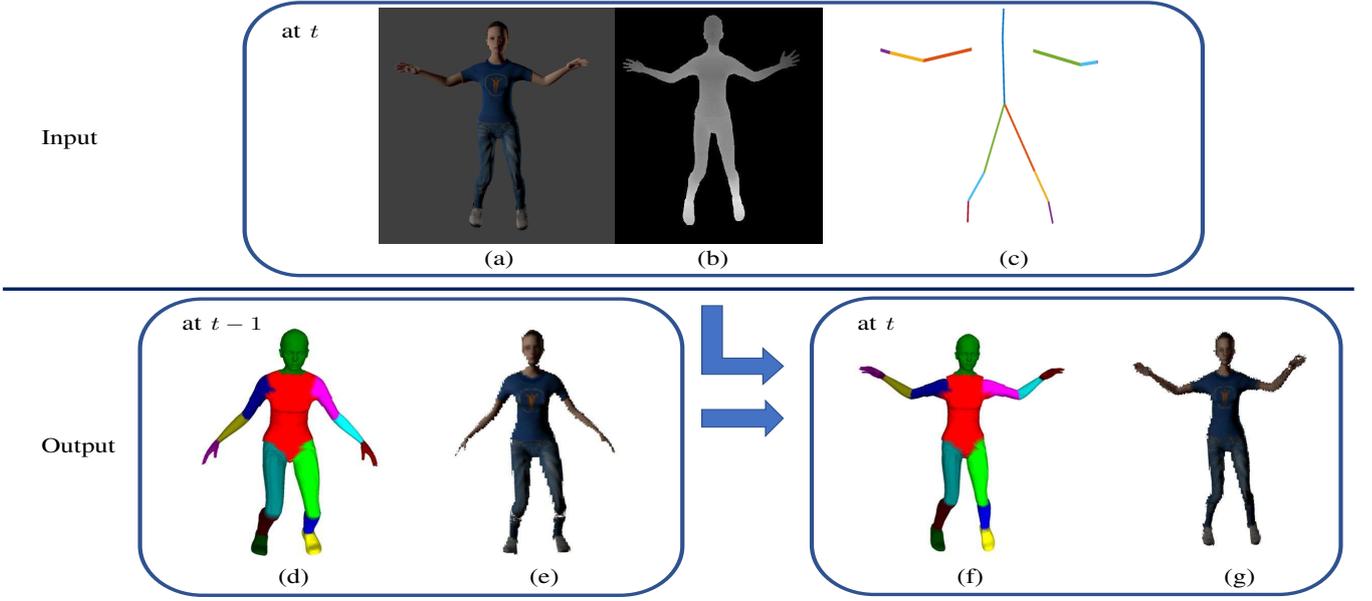
\begin{figure*}[ht]
\centering{
\def\svgwidth{\linewidth}
\resizebox{\linewidth}{8cm}{\input{figures//Methodology.pdf_tex}}
}
\caption{Block diagram of inputs and outputs of our proposed system. (a), (b) and (c) shows colour, depth and skeleton inputs at current frame. (d) and (e) illustrates puppet model and 3D reconstruction outputs at previous frame and finally (f) and (g) are puppet model and 3D reconstruction output at current frame. }
\label{fig:SystemPipeline} 
\end{figure*}

\section{Methodology}

Our approach operates in a frame-to-frame manner. For each frame, we sequentially perform three steps: at first, the motion is tracked by a puppet model using the current frame skeleton prior. Secondly, non-rigid tracking is carried out using the puppet model. Here the puppet model's transformations are used to initialize tracking and correspondence estimation. Thirdly, volumetric fusion is carried out as in state-of-the-art approaches \cite{yu122017bodyfusion, newcombe2015dynamicfusion, innmann2016volumedeform}. A block diagram of our proposed system is shown in Figure \ref{fig:SystemPipeline}. Note that unlike the other approaches \cite{yu122017bodyfusion,newcombe2015dynamicfusion} our proposed system takes the skeleton prior as an input per frame along with the RGB-D data.

\subsection{Motion Tracking using the Puppet Model}

In this step, we track the RGB-D input using a human model. Unlike other methods \cite{bogo2014faust,l0Norigid2015} which use the human model as a whole and use local optimization for tracking we treat the human model as a combination of rigid parts. Because of that, we term this model a \say{puppet model}. The skeleton joints at the current frame are used to initially align the puppet model with the incoming data. 

At first, the initial rigid transformation ($\mathbf{R}_{init}$, $t_{init}$) of each body part is calculated. Let a part has a bone with joint positions of $j_{head}^{t-1}$ and $j_{tail}^{t-1}$ in previous frame (at $t-1$) and corresponding joint positions in the current frame (at $t$) of $j_{head}^{t}$ and $j_{tail}^{t}$. $\mathbf{R}_{init}$ is estimated using angle between skeleton bones. Let $\vec{a}$ and $\vec{b}$ be unit vectors pass through skeleton bones as below,
\begin{equation} 
\vec{a}  = \frac{j_{head}^{t-1} -j_{tail}^{t-1}}{\left\Vert j_{head}^{t-1} -j_{tail}^{t-1} \right\Vert},\quad\quad 
\vec{b} = \frac{j_{head}^{t} - j_{tail}^{t}}{\left\Vert j_{head}^{t} - j_{tail}^{t} \right\Vert},
\end{equation}
and let's denote
\begin{equation} 
v = \vec{a} \times \vec{b}, \quad \quad c = \vec{a} \cdot \vec{b},
\end{equation}
\begin{equation} 
s = \left\Vert v \right\Vert.
\end{equation}
Then $\mathbf{R}_{init}$ is estimated as 
\begin{equation} 
\mathbf{R}_{init} = \mathds{1} + \left[ v \right]_{\times} + \left[ v \right]_{\times}^{2}\frac{1-c}{s^2} \label{rot_equation}
\end{equation}
where $\mathds{1}$ is the $3\times3$ identity matrix and $\left[ v \right]_{\times}$ is the skew-symmetric cross-product matrix of $v$ which defined as below,
\begin{equation} 
\left[ v \right]_{\times} \myeq   \begin{bmatrix}0 & -v_3 & v_2 \\
v_3 & 0 & -v_1 \\
-v_2 & v_1 & 0 \\
\end{bmatrix}.
\end{equation}
The Equation \ref{rot_equation} estimates the rotation matrix by using the angle between the skeleton bone vectors $\vec{a}$ and $\vec{b}$. After finding $\mathbf{R}_{init}$, $t_{init}$ is estimated as
\begin{equation} 
t_{init} = -\mathbf{R}_{init}c_1 + c_2,
\end{equation}
where $c_1$ and $c_2$ are mid points of the bones estimated as below,
\begin{equation} 
c_1 = \frac{j_{head}^{t-1} + j_{tail}^{t-1}}{2},  \quad c_2 = \frac{j_{head}^{t} + j_{tail}^{t}}{2}.
\end{equation}

After transforming each body part by the initial rigid transformation ($R_{init}$,$t_{init}$) three iterations of non-rigid ICP are carried out for aligning with the target cloud. Figure \ref{fig:SystemPipeline} (f) is the aligned puppet model after this step. We can see that through this method a quick initial alignment is possible whereas other approaches \cite{l0Norigid2015,yu122017bodyfusion} required many iterations for tracking.

\subsection{Non-rigid Tracking}

For each frame, the canonical model of the object is developed in SDF using the weighted average scheme of Curless and Levoy \cite{curless1996volumetric}. By integrating data from each frame, a warp field $\mathbf{W}$ is estimated for mapping points from canonical coordinates to live camera coordinates. $\mathbf{W}$ is modelled by a deformation graph \cite{sumner2007embedded} in which each graph node stores the rigid transformation matrices $\left\{\mathbf{T_i}\right\}$. The warp function $\mathbf{W}$, of a point $\mathbf{x}$ in canonical coordinate space is of the form shown below, 
\begin{equation}  
\mathbf{W}(\mathbf{x}) = \sum_{\mathbf{p_i} \in \mathcal{N}(\mathbf{x})} w(\mathbf{p_i}, \mathbf{x})\mathbf{T_i}[\mathbf{x^{T}} 1]^{T}, \label{position_tranfer} \end{equation}

$\mathbf{T_{rigid}}$ is $4\times4$ rigid transformation matrix, where $w(\mathbf{p_i}, \mathbf{x})$ is the influence weight of point $\mathbf{x}$ for node $i$; $\mathcal{N}(\mathbf{x})$ is the set of nearest graph nodes from $\mathbf{x}$ and $\mathbf{p_i}$ is the position of the $i^\text{th}$ graph node. In this step, we are using the aligned puppet model for two purposes: for initializing tracking and correspondence estimation. Similarly, the normal $\mathbf{n}$ at a point $\mathbf{x}$ is transformed by the warp field $\mathbf{W}$,
\begin{equation} \mathbf{W}(\mathbf{n}) = \sum_{\mathbf{p_i} \in \mathcal{N}(\mathbf{x})} w(\mathbf{p_i}, \mathbf{x})\mathbf{T_i}[\mathbf{n}^{T} 0]^{T}.  \label{normal_tranfer} \end{equation}

\subsubsection{Initializing Tracking} \label{deformation_graph} Let the warp function corresponding to previous frame be $\mathbf{W_{t-1}}$. To estimate the warp function at the current frame, $\mathbf{W_{t}}$ ,we first initialize it as, 
\begin{equation}
\mathbf{W_{t}^{*}} = \mathbf{W_{t}^{p}W_{t-1}},
\end{equation}
where $W_{t}^{p}$ is the warp field imposed by the puppet model at the current frame which is estimated as, 
\begin{equation}
\mathbf{W_{t}^{p}}(\mathbf{x}) =  \sum_{\mathbf{q_k} \in \mathcal{N}(\mathbf{x})} w(\mathbf{q_k}, \mathbf{x}) \mathbf{T_k},
\end{equation}
where $\mathbf{T_k}$ is the $4\times4$ rigid transformation matrix $w(\mathbf{q_k}, \mathbf{x})$ is the influence weight of point $\mathbf{x}$;
$\mathcal{N}(\mathbf{x})$ is the set of nearest neighbours of $\mathbf{x}$ in the puppet model and $\mathbf{q_k}$ is the position of the $k^{th}$ vertex in the puppet model. Here the transformation ($\mathbf{T_k}$) is calculated as,
\begin{equation}
\mathbf{T_{k}} = \sum_{j}w_{j}\mathbf{T_j},
\end{equation}
where $w_j$ is the skinning weight corresponding to the $j^{th}$ bone and $\mathbf{T_{j}}$ is the rigid transformation estimated for the $j^{th}$ bone of the skeleton.  

\begin{figure}[t]
\centering{
\def\svgwidth{\linewidth}
\resizebox{\linewidth}{4.5cm}{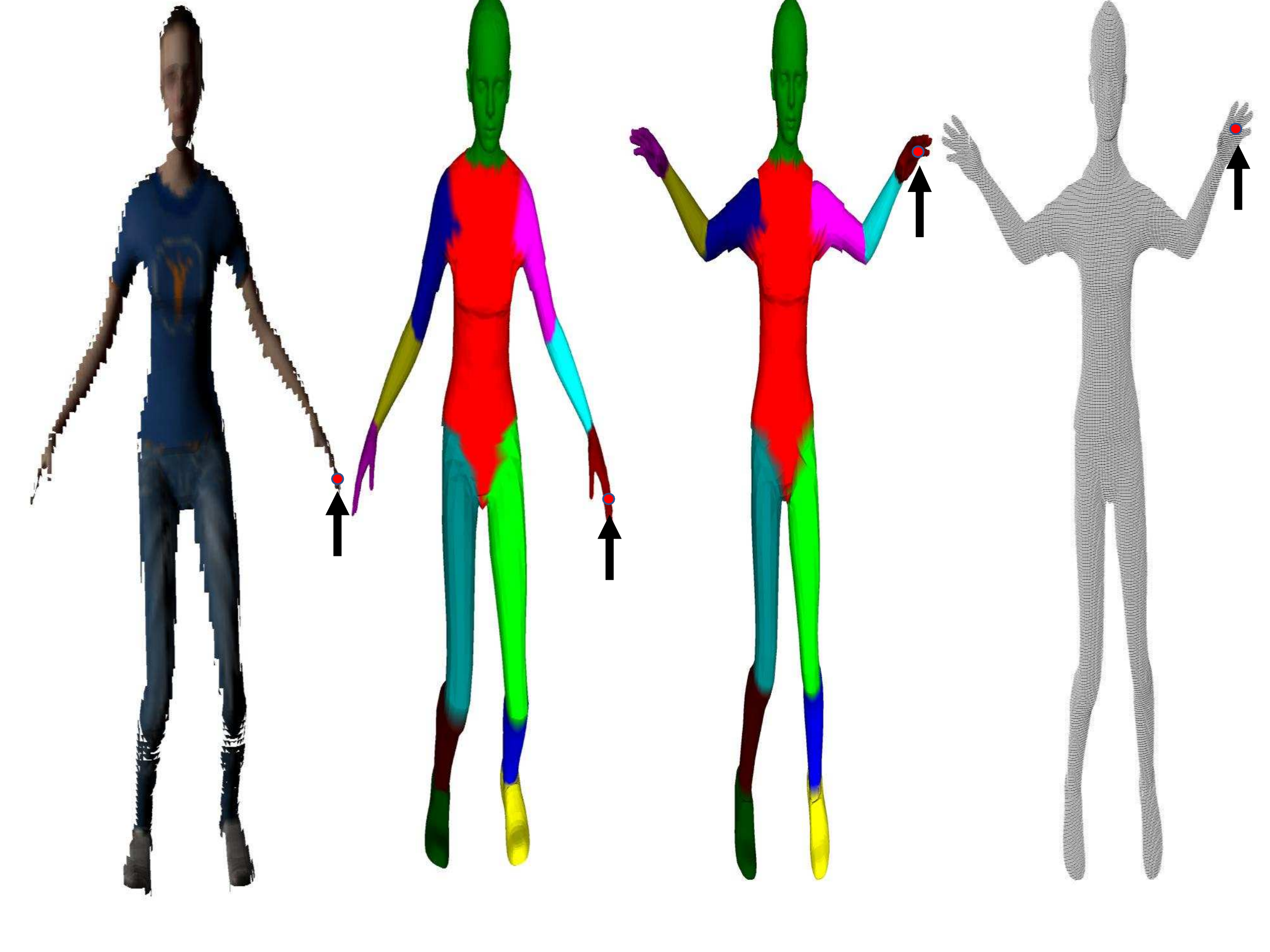}
}

\caption{Illustration of the correspondence estimation approach. For a point in the reconstruction $v_c$ the nearest neighbour in the puppet model $v_c^p$ is estimated (shown in first and second images from left to right). The corresponding point of $v_c^p$ in the aligned puppet $v_l^p$ is used for finding nearest neighbour in the target cloud $v_t$ (shown in third and forth images). The correspondence from $v_c$ to $v_t$ is established in this way (refer to Section \ref{CorrEst}).}
\label{fig:Correspondence Estimation} 
\end{figure}

\subsubsection{Correspondence Estimation} \label{CorrEst} For each point in the reconstruction $v_c$, the nearest neighbor is found $v_c^p$ from the puppet model.
The corresponding point of $v_c^p$ in the aligned puppet $v_l^p$ is used for finding nearest neighbour in the target cloud $v_t$. In this way, we establish correspondences to the target cloud through the puppet model. Figure \ref{fig:Correspondence Estimation} illustrates this correspondence estimation approach. A projective data association is used for finding the nearest neighbor. The correspondences through the joint regions of the puppet model is omitted since joint regions usually have complex deformations. Therefore using them often leads to tracking failure. Note that this kind of correspondence is used just for the first iteration after that correspondences are estimated from the warped mesh to the target cloud directly.

We track non-rigid human subjects based on two types of constraints: deformation graph and skeleton. In the deformation graph, the non-rigidity is modeled by a sub-sampled graph as explained in \ref{deformation_graph}. The skeleton is used for quickly tracking articulated motions of human subjects. Similar to the well known Inverse Kinematics (IK) problem, we model the human body as a set of joints. We model hips as the root joint having 6 DoFs, and all the other joints having 3 DoFs (rotational only).

The energy function for our non rigid tracking is given below,
\begin{multline} \label{energy function} 
E_{non-rigid} = \alpha_{data}E_{data} + \alpha_{arap}E_{arap} + \alpha_{skeleton}E_{skeleton} \\ 
+ \alpha_{reg}E_{reg},  
\end{multline}
where $E_{data}$ is the error driven by the deformation graph constraints, $E_{arap}$ is the local as-rigid-as-possible constraint imposed on neighbouring graph nodes, $E_{skeleton}$ is the error driven by the skeleton constrains and $E_{reg}$ is the regularization term for encouraging skeleton and deformation constraints to agree with each other. 
$E_{data}$  and $E_{arap }$ are defined as follows:
\begin{equation}
E_{data} = \sum_{(v_i,u_i)\in \boldsymbol C}\left\vert \tilde n_{v_i}( \tilde v_i - u_i) \right\vert^2, \label{data} \ 
\end{equation}
\begin{equation}
E_{arap} = \sum_{i}\sum_{ j \in \mathcal{N}(i)}\left\vert(v_i - v_j) - \boldsymbol R_i(\tilde v_i - \tilde v_j) \right\vert^2, \label{ARAP} 
\end{equation}
where $\boldsymbol C$ represents the set of correspondences, $\tilde v_i$, and $ \tilde n_i$ represents vertex coordinates and its normal warped by the deformation graph, and $\mathcal{N}(i)$ represents the nearest neighbor set of the $i^{th}$node in the graph. 

Similar to $E_{data}$ (Equation \ref{data}) $E_{skeleton}$ also represents point to plane error as defined below,
\begin{equation}
E_{skeleton} = \sum_{(v_i,u_i)\in C}\left\vert \hat n_{v_i}( \hat v_i - u_i) \right\vert^2, \label{skeleton} \ 
\end{equation}
where $\hat n_{v_i}$ and $\hat v_i$ are the normal and vertex coordinates warped by the skeleton defined as
\begin{equation}
\hat v_{i} = \sum_{j \in \mathcal{B}} w_{i,j} \boldsymbol T_{bj}v_{i}, \quad \quad \hat n_{v_i} = \sum_{j \in \mathcal{B}} w_{i,j} \boldsymbol T_{bj}n_{v_i}, \label{skeleton_warp} \ 
\end{equation}
where $\boldsymbol T_{bj}$ is the deformation associated with the $j^{th}$ bone of the skeleton. $w_{i,j}$ is the skinning weight for the $j^{th}$ bone. The skeleton is modeled as the kinematic chain as explained in \cite{buss2004introduction}. The skinning weights each graph node is taken as the skinning weights of nearest neighbour from the puppet model. The last term in our energy function (Equation \ref{energy function}), $E_{reg}$, enforces the non-rigid deformation modeled by both the deformation graph and the skeleton to agree with each other, 
\begin{equation}
E_{reg} = \sum_{i=1}^{N}\left\vert \tilde v_{i} -\hat v_i \right\vert^2, \label{regularization} \ 
\end{equation}
where $N$ is the number of nodes in the deformation graph. The skeleton helps to tracks articulated motions quickly and the deformation graph helps to model the non-rigid surface. The role of $E_{reg}$ is to connect deformation  graph  and  the  skeleton and take advantage of both approaches. The Equation \ref{energy function} is a non-linear least squares problem. We solve this using a GPU-based Gauss-Newton solver similar to \cite{zollhofer2014real}.

%% file: figures/Methodology.pdf_tex
\begingroup%
  \makeatletter%
  \providecommand\color[2][]{%
    \errmessage{(Inkscape) Color is used for the text in Inkscape, but the package 'color.sty' is not loaded}%
    \renewcommand\color[2][]{}%
  }%
  \providecommand\transparent[1]{%
    \errmessage{(Inkscape) Transparency is used (non-zero) for the text in Inkscape, but the package 'transparent.sty' is not loaded}%
    \renewcommand\transparent[1]{}%
  }%
  \providecommand\rotatebox[2]{#2}%
  \newcommand*\fsize{\dimexpr\f@size pt\relax}%
  \newcommand*\lineheight[1]{\fontsize{\fsize}{#1\fsize}\selectfont}%
  \ifx\svgwidth\undefined%
    \setlength{\unitlength}{960bp}%
    \ifx\svgscale\undefined%
      \relax%
    \else%
      \setlength{\unitlength}{\unitlength * \real{\svgscale}}%
    \fi%
  \else%
    \setlength{\unitlength}{\svgwidth}%
  \fi%
  \global\let\svgwidth\undefined%
  \global\let\svgscale\undefined%
  \makeatother%
  \begin{picture}(1,0.5625)%
    \lineheight{1}%
    \setlength\tabcolsep{0pt}%
    \put(0,0){\includegraphics[width=\unitlength,page=1]{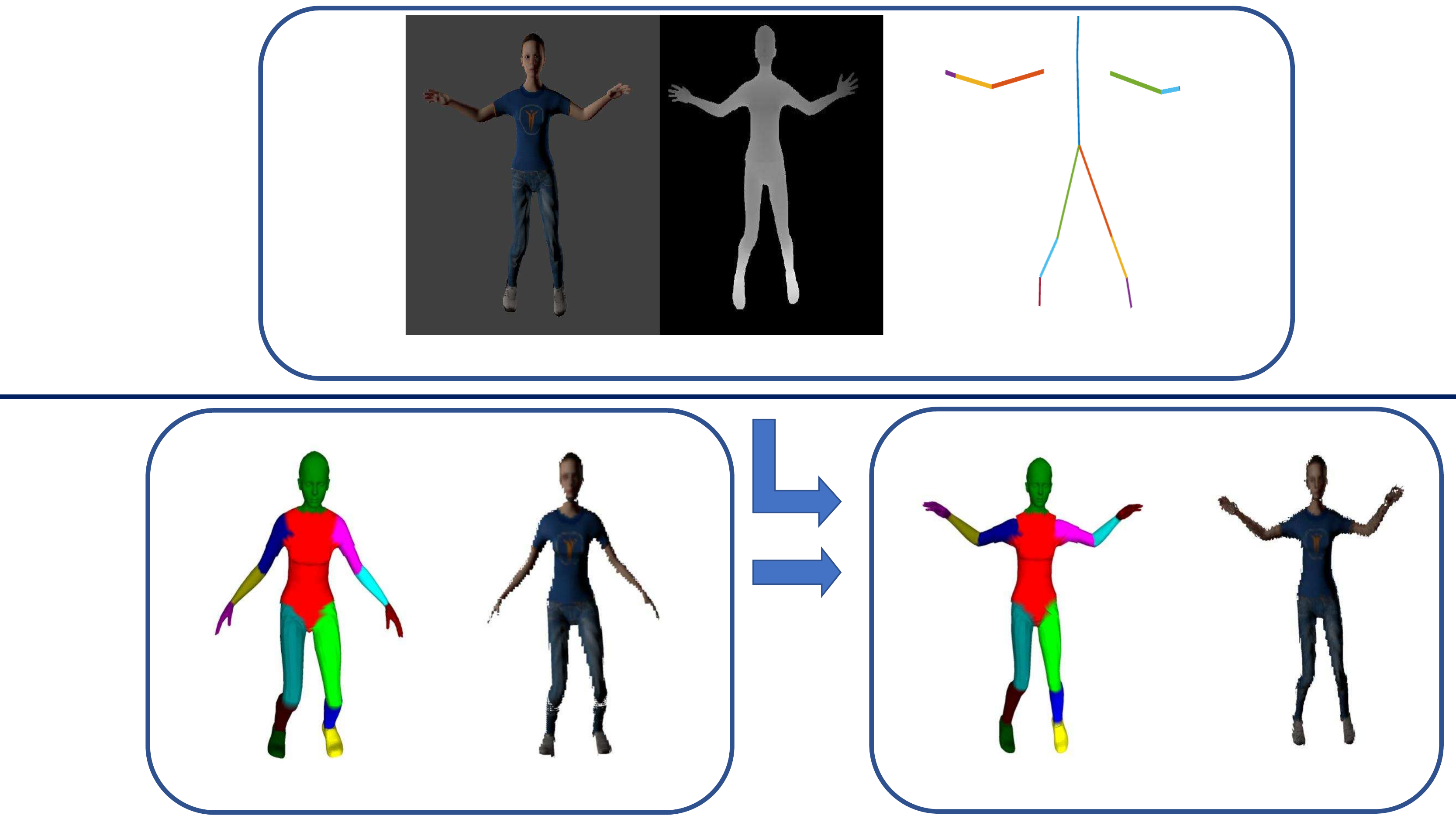}}%
    \put(0.35642092,0.31222331){\color[rgb]{0,0,0}\makebox(0,0)[lt]{\lineheight{1.25}\smash{\begin{tabular}[t]{l}\fontsize{9pt}{1em}(a)\end{tabular}}}}%
    \put(0.51411269,0.31222331){\color[rgb]{0,0,0}\makebox(0,0)[lt]{\lineheight{1.25}\smash{\begin{tabular}[t]{l}\fontsize{9pt}{1em}(b)\end{tabular}}}}%
    \put(0.73051922,0.31222331){\color[rgb]{0,0,0}\makebox(0,0)[lt]{\lineheight{1.25}\smash{\begin{tabular}[t]{l}\fontsize{9pt}{1em} (c)\end{tabular}}}}%
    \put(0.19692523,0.01612956){\color[rgb]{0,0,0}\makebox(0,0)[lt]{\lineheight{1.25}\smash{\begin{tabular}[t]{l}\fontsize{9pt}{1em} (d)\end{tabular}}}}%
    \put(0.38286272,0.01612956){\color[rgb]{0,0,0}\makebox(0,0)[lt]{\lineheight{1.25}\smash{\begin{tabular}[t]{l}\fontsize{9pt}{1em} (e)\end{tabular}}}}%
    \put(0.69926928,0.01612956){\color[rgb]{0,0,0}\makebox(0,0)[lt]{\lineheight{1.25}\smash{\begin{tabular}[t]{l}\fontsize{9pt}{1em} (f)\end{tabular}}}}%
    \put(0.89692557,0.01617647){\color[rgb]{0,0,0}\makebox(0,0)[lt]{\lineheight{1.25}\smash{\begin{tabular}[t]{l}\fontsize{9pt}{1em} (g)\end{tabular}}}}%
    \put(0.02165559,0.42485308){\color[rgb]{0,0,0}\makebox(0,0)[lt]{\lineheight{1.25}\smash{\begin{tabular}[t]{l}\fontsize{9pt}{1em} Input\end{tabular}}}}%
    \put(0.02165559,0.13764101){\color[rgb]{0,0,0}\makebox(0,0)[lt]{\lineheight{1.25}\smash{\begin{tabular}[t]{l}\fontsize{9pt}{1em} Output\end{tabular}}}}%
    \put(0.19936773,0.52398471){\color[rgb]{0,0,0}\makebox(0,0)[lt]{\lineheight{1.25}\smash{\begin{tabular}[t]{l}\fontsize{9pt}{1em} at $t$\end{tabular}}}}%
    \put(0.12587301,0.25245771){\color[rgb]{0,0,0}\makebox(0,0)[lt]{\lineheight{1.25}\smash{\begin{tabular}[t]{l}\fontsize{9pt}{1em} at $t-1$\end{tabular}}}}%
    \put(0.62549371,0.25245771){\color[rgb]{0,0,0}\makebox(0,0)[lt]{\lineheight{1.25}\smash{\begin{tabular}[t]{l}\fontsize{9pt}{1em} at $t$\end{tabular}}}}%
  \end{picture}%
\endgroup%

%% file: figures/CorrespondenceEstimation.pdf_tex
\begingroup%
  \makeatletter%
  \providecommand\color[2][]{%
    \errmessage{(Inkscape) Color is used for the text in Inkscape, but the package 'color.sty' is not loaded}%
    \renewcommand\color[2][]{}%
  }%
  \providecommand\transparent[1]{%
    \errmessage{(Inkscape) Transparency is used (non-zero) for the text in Inkscape, but the package 'transparent.sty' is not loaded}%
    \renewcommand\transparent[1]{}%
  }%
  \providecommand\rotatebox[2]{#2}%
  \newcommand*\fsize{\dimexpr\f@size pt\relax}%
  \newcommand*\lineheight[1]{\fontsize{\fsize}{#1\fsize}\selectfont}%
  \ifx\svgwidth\undefined%
    \setlength{\unitlength}{720bp}%
    \ifx\svgscale\undefined%
      \relax%
    \else%
      \setlength{\unitlength}{\unitlength * \real{\svgscale}}%
    \fi%
  \else%
    \setlength{\unitlength}{\svgwidth}%
  \fi%
  \global\let\svgwidth\undefined%
  \global\let\svgscale\undefined%
  \makeatother%
  \begin{picture}(1,0.75)%
    \lineheight{1}%
    \setlength\tabcolsep{0pt}%
    \put(0,0){\includegraphics[width=\unitlength,page=1]{CorrespondenceEstimation.pdf}}%
    \put(0.25110864,0.26473521){\color[rgb]{0,0,0}\makebox(0,0)[lt]{\lineheight{1.25}\smash{\begin{tabular}[t]{l}\fontsize{12pt}{1em} $v_c$\end{tabular}}}}%
    \put(0.47300555,0.25640192){\color[rgb]{0,0,0}\makebox(0,0)[lt]{\lineheight{1.25}\smash{\begin{tabular}[t]{l}\fontsize{12pt}{1em} $v_c^p$\end{tabular}}}}%
    \put(0.70954525,0.53817273){\color[rgb]{0,0,0}\makebox(0,0)[lt]{\lineheight{1.25}\smash{\begin{tabular}[t]{l}\fontsize{12pt}{1em} $v_l^p$\end{tabular}}}}%
    \put(0.95226539,0.56213106){\color[rgb]{0,0,0}\makebox(0,0)[lt]{\lineheight{1.25}\smash{\begin{tabular}[t]{l}\fontsize{12pt}{1em} $v_t$\end{tabular}}}}%
  \end{picture}%
\endgroup%

%% file: sections/synthetic_data.tex
\section{synthetic data} \label{syntheticData}
There are only a few publicly datasets available for evaluating RGB-D based non-rigid 3D reconstruction. Those datasets \cite{innmann2016volumedeform,slavcheva2017killingfusion} are for general non-rigid subjects and not specific to humans. Even though the dataset published with \cite{elanattil2018non} has the frame-to-frame live ground truth geometry and camera trajectory~\cite{elanattil2018syn}, they do not have ground truth skeleton joints. We found that skeleton joint detection play an important role in human performance capture algorithms. Motivated by this we developed a synthetic dataset which has ground truth for frame-to-frame geometry and skeleton joint detection. 

\begin{figure}[t!]
    \includegraphics[width =\linewidth,height=4cm]{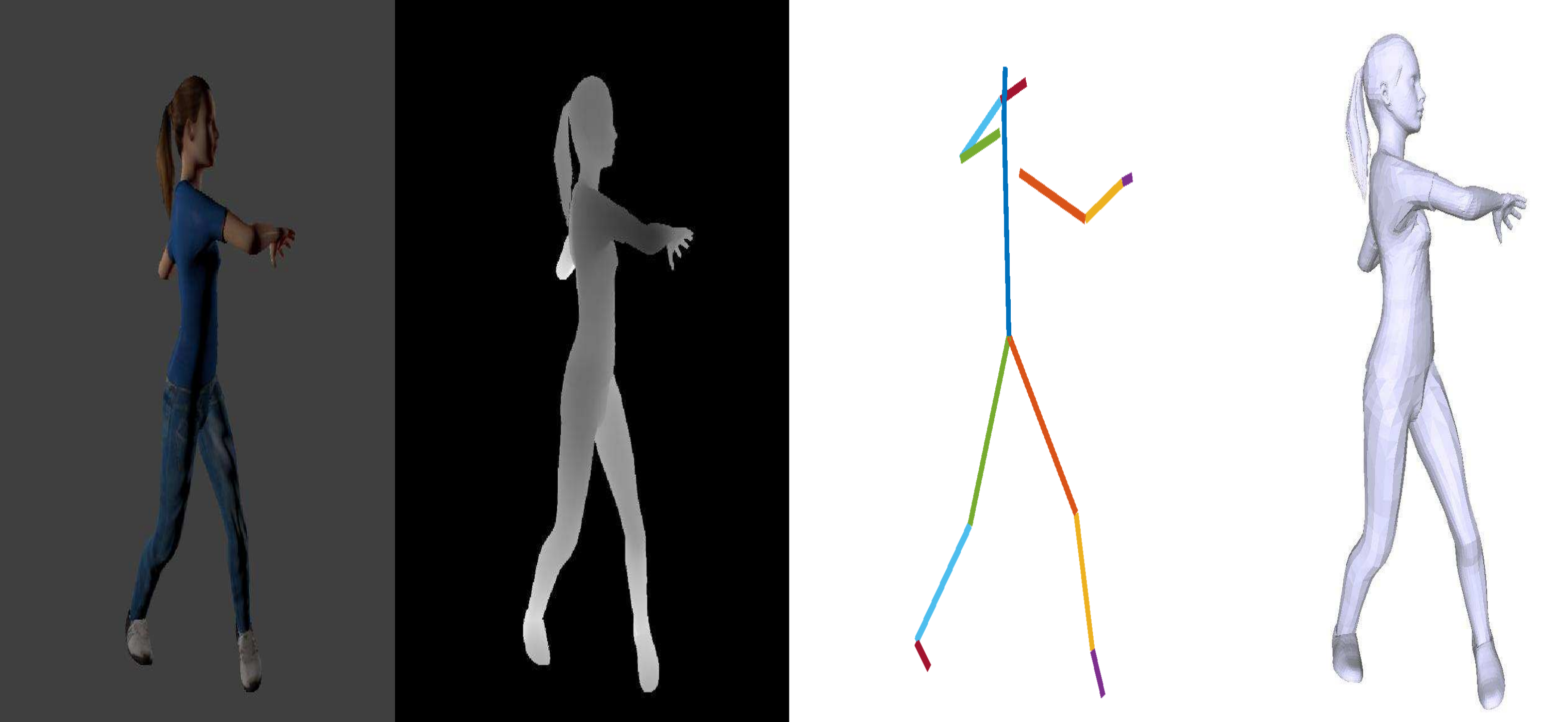}
    \caption{Each frame in the synthetic data consists of RGB image, depth image, skeleton and ground truth geometry (shown from left to right). }
\label{fig:SyntheticData} 
\end{figure}

\begin{figure}[]
\begin{center}
\begin{tabular}{llllll}
\hline  \noalign{\smallskip}
\multicolumn{1}{c}{Name} & \multicolumn{1}{c}{$N$} & Mean   & \multicolumn{1}{c}{Min} & Max    & \multicolumn{1}{c}{Std} \\ \hline \noalign{\smallskip} \noalign{\smallskip}
Jump Balance               & 60                     & 0.988 & 0.263                   & 2.605 & 0.629                   \\ \noalign{\smallskip}
Punch Strike               & 250                    & 0.444 & 0.084                   & 0.938 & 0.201                   \\ \noalign{\smallskip}
Boxing                     & 245                    & 0.650 & 0.015                   & 1.589 & 0.312                   \\ \noalign{\smallskip}
Sword Play                 & 248                    & 0.521 & 0.082                   & 1.165 & 0.252                   \\ \noalign{\smallskip}
Exercise                   & 248                    & 0.733  & 0.068                   & 1.919 & 0.456                    \\ \noalign{\smallskip}
Kick Ball                  & 161                    & 0.536 & 0.030                   & 2.752 & 0.607                   \\ \noalign{\smallskip}
Direct Traffic             & 251                    & 0.578 & 0.126                   & 1.912 & 0.260                   \\ \noalign{\smallskip}\noalign{\smallskip}\hline
\end{tabular}
\captionof{table}{Details of the synthetic data. Each row have sequence name, number of frames in sequence ($N$), and statistics of joint motion are given (refer section \ref{syntheticData}).  \label{tbl:SyntheticData}}\end{center}
\end{figure}
For generating synthetic data, at first, we create a human body model using the MakeHuman package\footnote{MakeHuman \href{http://www.makehumancommunity.org}{http://www.makehumancommunity.org}}. MakeHuman is an open source tool for making 3D human characters. Then we export this character to Blender\footnote{Blender \href{https://www.blender.org }{https://www.blender.org}}. 
Then we use the CMU motion capture dataset\footnote{Motionbuilder-friendly BVH conversion \href{https://sites.google.com/a/cgspeed.com/cgspeed/motion-capture/cmu-bvh-conversion}{CMU's Motion Capture Database}} to animate the 3D model. This enables us to simulate a wide variety of human movements with skeleton joint detection. Each frame in our synthetic dataset consists of an RGB image, depth image, skeleton and ground truth geometry. Figure \ref{fig:SyntheticData} shows this data in a frame from the \textquoteleft exercise\textquoteright \hspace{1pt} dataset. We have made our database publicly available to enable researchers to replicate our results and contribute to further advances in the area\footnote{Our Synthetic data is publicly available at \href{https://research.csiro.au/robotics/databases/}{https://research.csiro.au/ \linebreak robotics/databases} or \href{https://research.qut.edu.au/saivt/}{https://research.qut.edu.au/saivt/}}.

We developed seven data sequences of varying motions characteristics. Table \ref{tbl:SyntheticData} shows motion statistics of the corresponding data sequences. The motion is estimated as the sum of joint movement in each frame. We assign the same name as used in the CMU Mocap dataset for each sequence. The first two columns in Table \ref{tbl:SyntheticData} show name and number of frames in the sequence. The remaining columns shows the motion statistics for each data sequence. Note that for the current work we develop data just for a single subject with a static camera. However our framework is capable for making data for multiple subjects with given camera trajectory similar to \cite{elanattil2018non}.

%% file: sections/experiments.tex
\section{Experiments}
In this section, we describe  the  qualitative  and  quantitative evaluation  of  our reconstruction framework. For qualitative evaluation of our reconstruction, we estimate Mean Absolute Error (MAE) of point to plane distances from each point to ground-truth geometry.
\begin{figure}
\begin{center}
\begin{tabular}{cccccc}\hline\noalign{\smallskip}
 &  Iteration & Mean & Std. & Hausdroff & Outliers \\
\hline\noalign{\smallskip}
\multirow{4}{4em}{case 1} & 1  & 16.1 & 15.3 & 0.2155 & 7599\\
& 2 & 11.2  & 11.6 & 0.2095  & 6464\\
& 3 & \hspace{0.18cm}9.4  & 11.0 & 0.2088  & 5451   \\
& 4 & \hspace{0.18cm}8.6  & 10.3 & 0.2070  & 5027 \\ 
\hline\noalign{\smallskip}
\multirow{4}{4em}{case 2} & 1  & 13.8 & 13.5 & 0.2107 & 7099\\
& 2 & 10.2  & 10.6 & 0.2026  & 5913\\
& 3 & \hspace{0.18cm}8.5  & \hspace{0.18cm}9.6 & 0.2013  & 5179\\
& 4 & \hspace{0.18cm}7.7  & \hspace{0.18cm}8.6 & 0.1978  & 4840 \\ \hline
\end{tabular}
\captionof{table}{Qualitative comparison between (case1) tracking without any initialization and (case2) tracking with initialization using puppet model's rigid transformation (refer section \ref{PuppetBasedInitialization}). Also see Figure \ref{fig:QualitativeComparison} for reconstruction results in both cases. Here the unit of mean and standard deviation is mm. \label{tbl:trackingFromPuppetModel}}
\end{center}
\end{figure}

\begin{figure}[t]
\begin{center}
\includegraphics[width=\linewidth, height = 4cm]{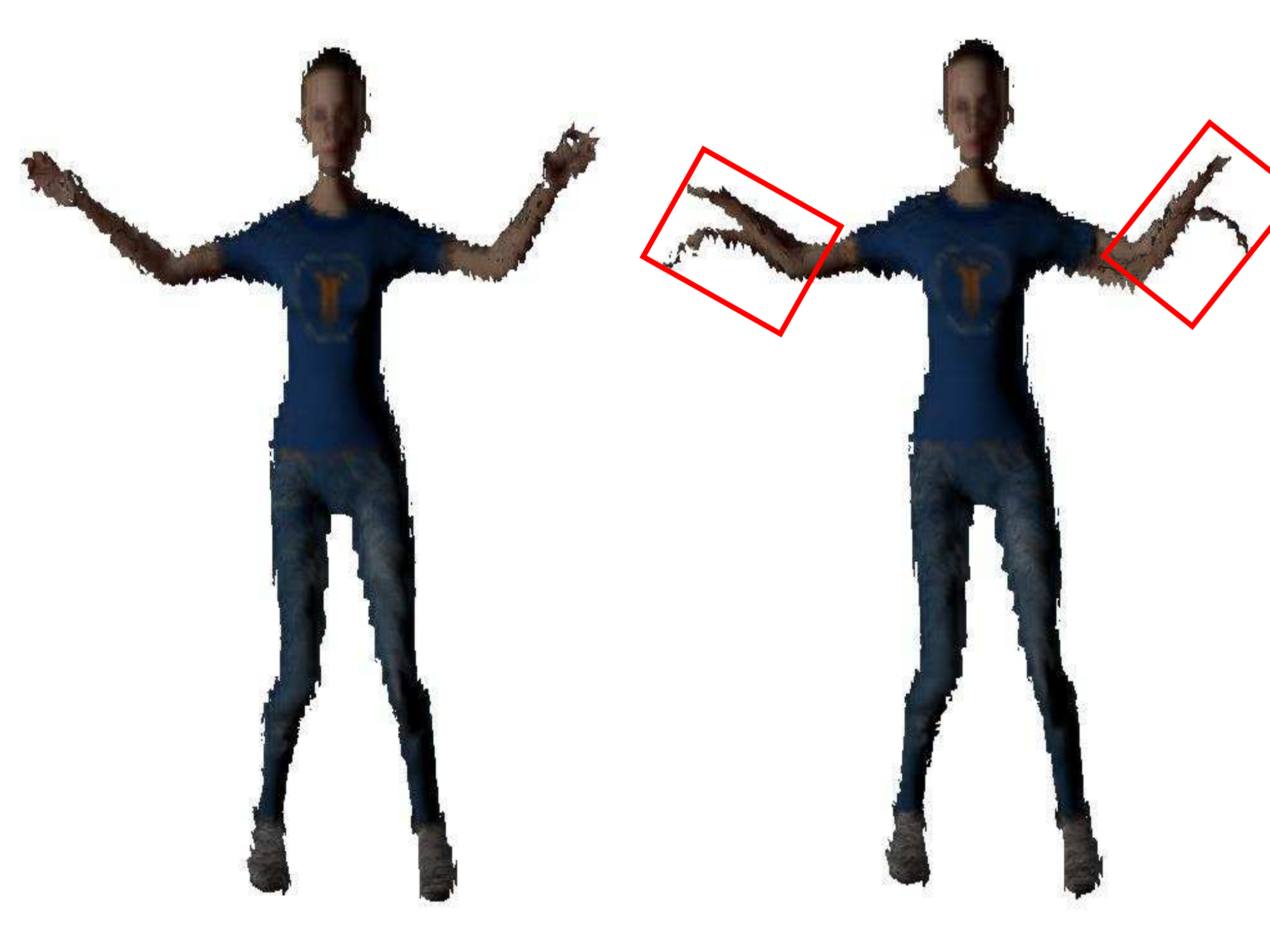}
\end{center}

\caption{Advantages of using the the puppet model's rigid transformations to initialize tracking. Left with initialization and right without initialization. We can see that in the right image tracking failed and error accumulated in the 3D reconstruction.}
\label{fig:QualitativeComparison} 
\end{figure}
\begin{figure}[!htb]
\begin{center}
   \includegraphics[width=\linewidth]{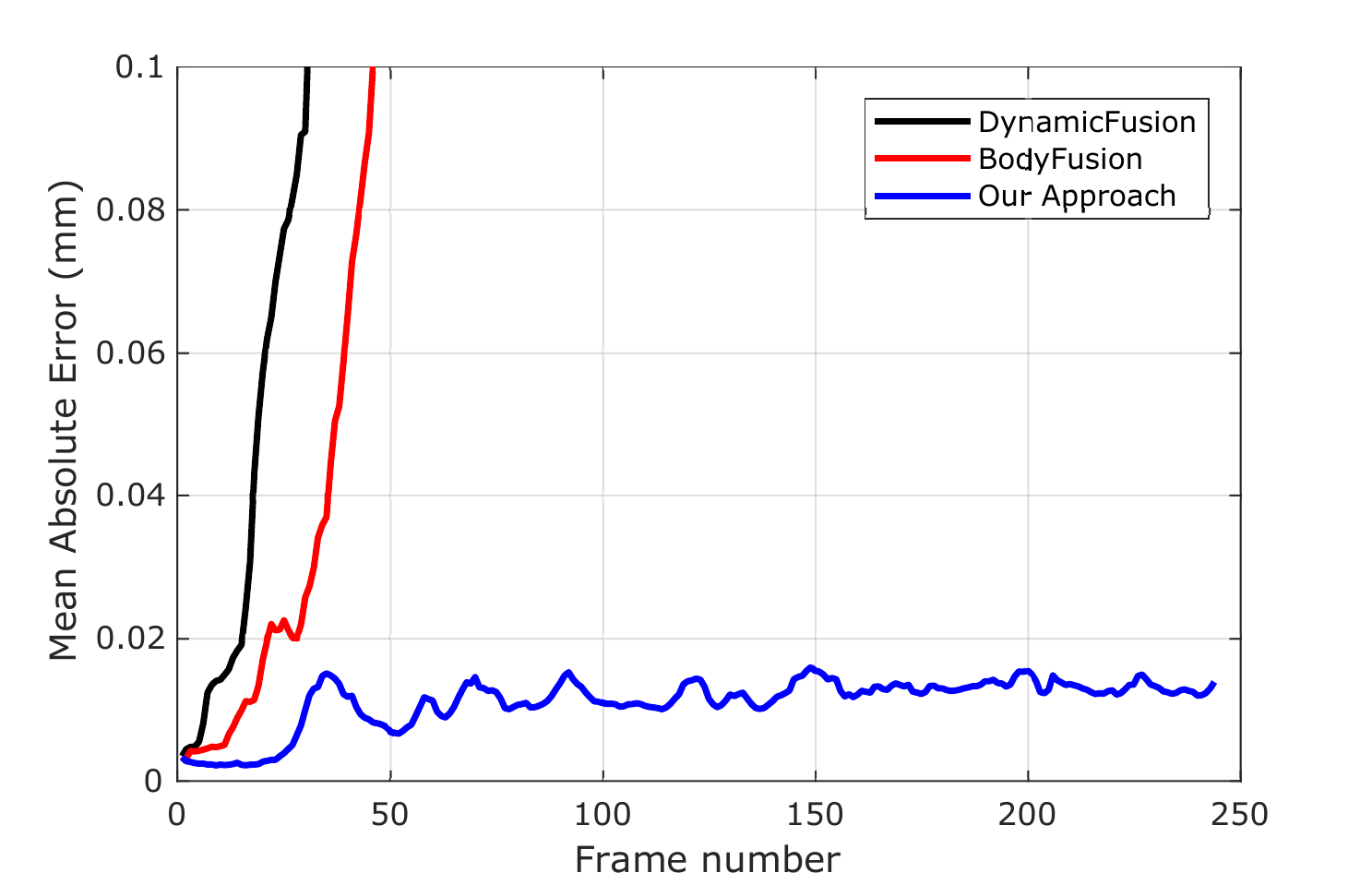}
\end{center}
 \caption{Average error on each frame is plotted from the \textquoteleft Boxing\textquoteright\hspace{1pt} data. Black and red correspond to using DynamicFusion \cite{newcombe2015dynamicfusion} and BodyFusion \cite{yu122017bodyfusion} respectively and blue corresponds to our method.}
 \label{Quant_Boxing}
\end{figure}
\begin{figure}[!htb]
\begin{center}
   \includegraphics[width=\linewidth]{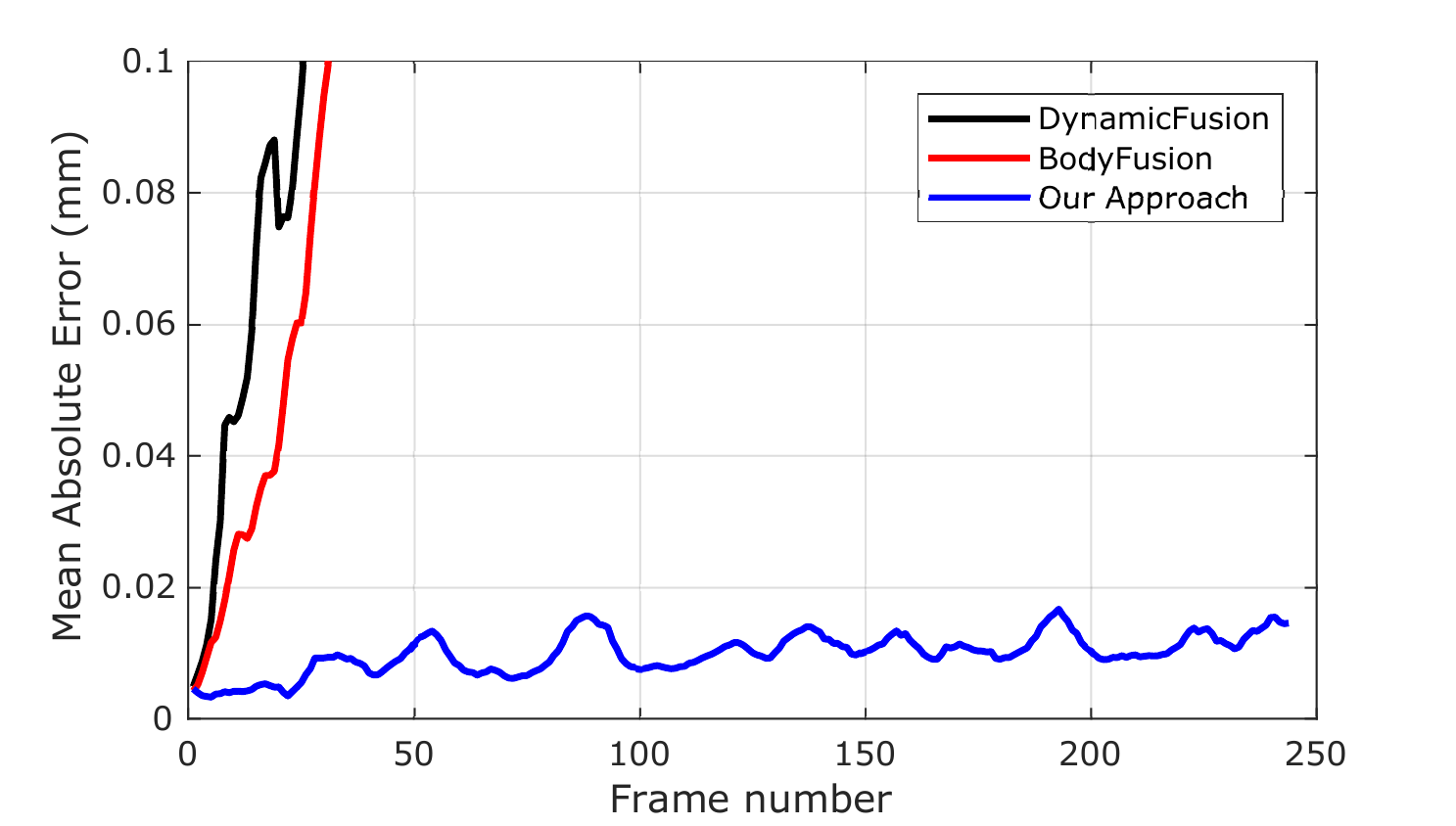}
\end{center}
 \caption{Average error on each frame is plotted from the \textquoteleft Punch Strike\textquoteright\hspace{1pt} data. Black and red correspond to using DynamicFusion \cite{newcombe2015dynamicfusion} and BodyFusion \cite{yu122017bodyfusion} respectively and blue corresponds to our method.}
 \label{Quant_PunchStrike}
\end{figure}
\begin{figure}[!htb]
\begin{center}
   \includegraphics[width=\linewidth, height =8cm]{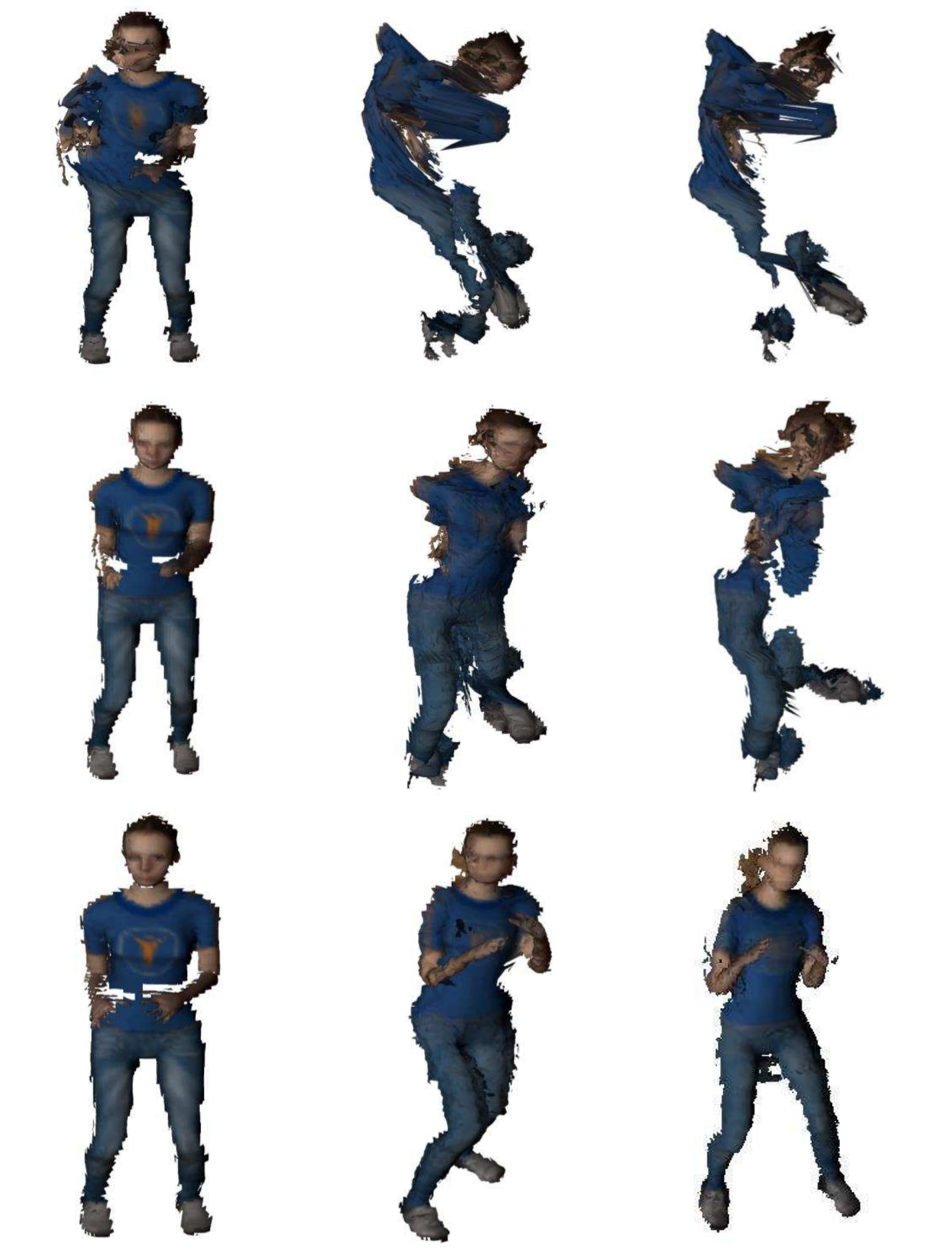}
\end{center}
 \caption{Qualitative comparison with state-of-the-art approaches from the \textquoteleft Boxing\textquoteright\hspace{1pt} data. The top row shows reconstruction from DynamicFusion \cite{newcombe2015dynamicfusion} and the middle row shows reconstruction from BodyFusion \cite{yu122017bodyfusion} and the bottom row is the reconstruction from our approach.}
 \label{Qual_Boxing}
\end{figure}

\begin{figure*}[!htb]
    \begin{center}
    \stackunder{\includegraphics[width= 0.16\linewidth,height=7cm]{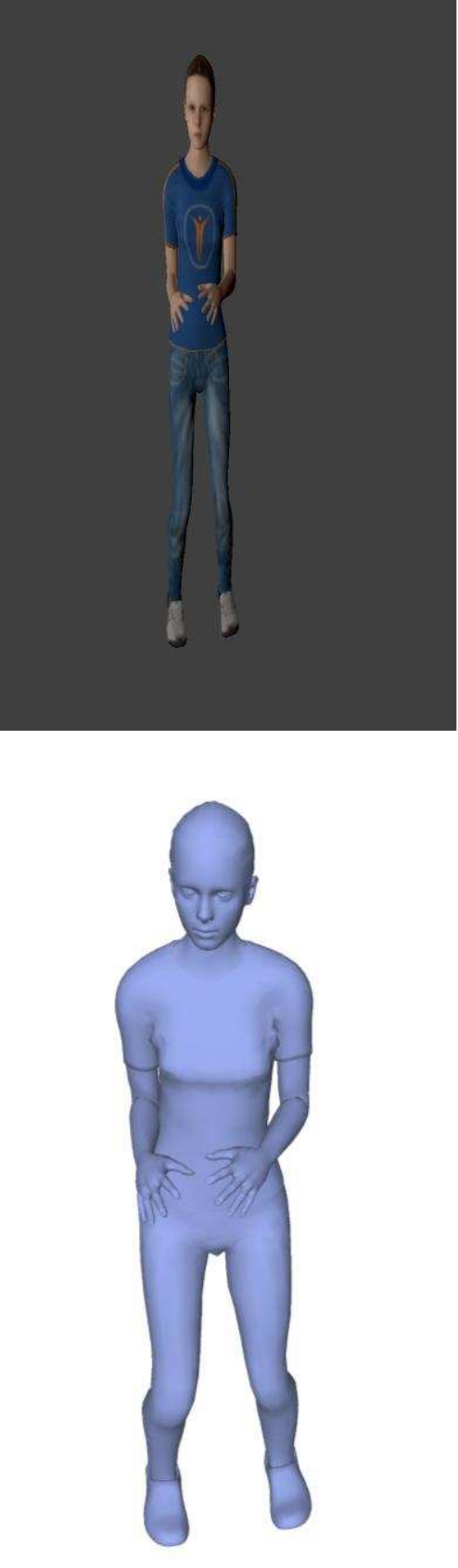}}{\#1}\hspace{0.5pt}
    \stackunder{\includegraphics[width= 0.16\linewidth,height=7cm]{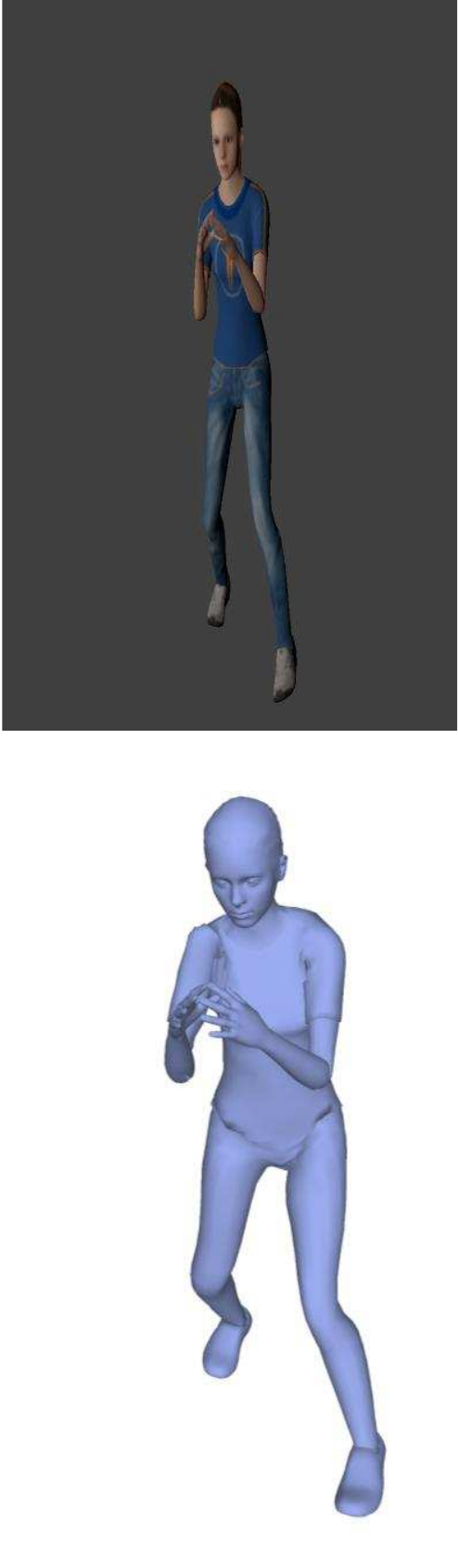}}{\#42}\hspace{0.5pt}
    \stackunder{\includegraphics[width= 0.16\linewidth,height=7cm]{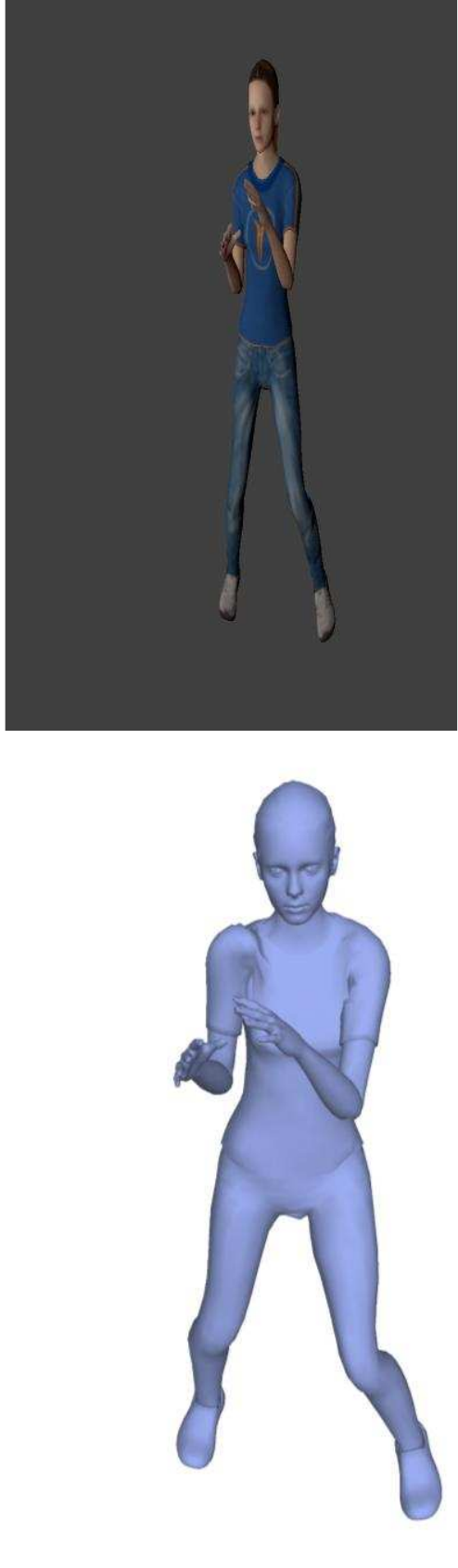}}{\#88}\hspace{0.5pt}
    \stackunder{\includegraphics[width= 0.16\linewidth,height=7cm]{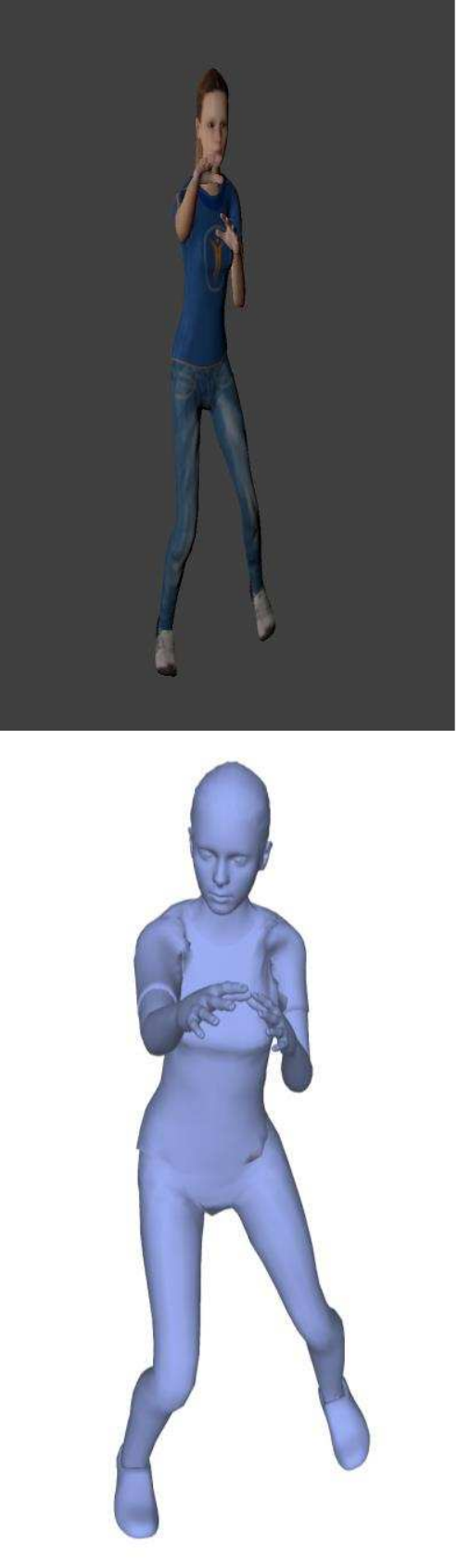}}{\#130}\hspace{0.5pt}
    \stackunder{\includegraphics[width= 0.16\linewidth,height=7cm]{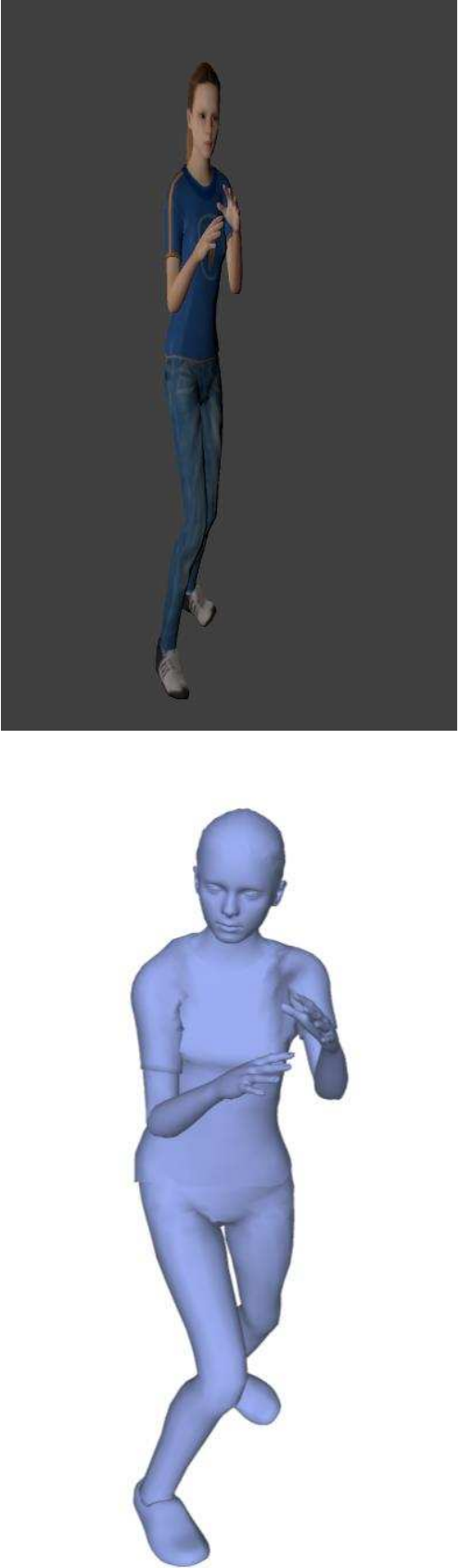}}{\#194}\hspace{0.5pt}
    \stackunder{\includegraphics[width= 0.16\linewidth,height=7cm]{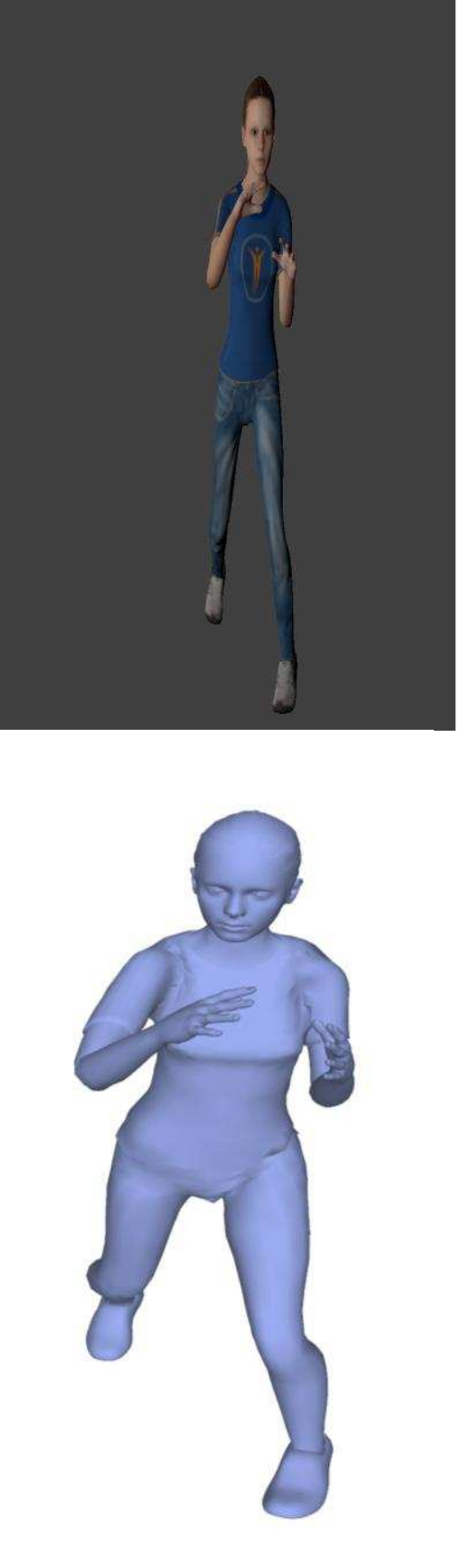}}{\#219}
    \end{center}
    \caption{Qualitative results of motion tracking from the \textquoteleft Boxing\textquoteright \hspace{1pt} sequence. The upper row shows images of different frames and the lower row shows the respective deformed 3D model. The frame index is shown below each image.}
    \label{QualitativeBoxing}
\end{figure*}

\begin{figure*}[!htb]
    \begin{center}
    \stackunder{\includegraphics[width= 0.16\linewidth,height=7cm]{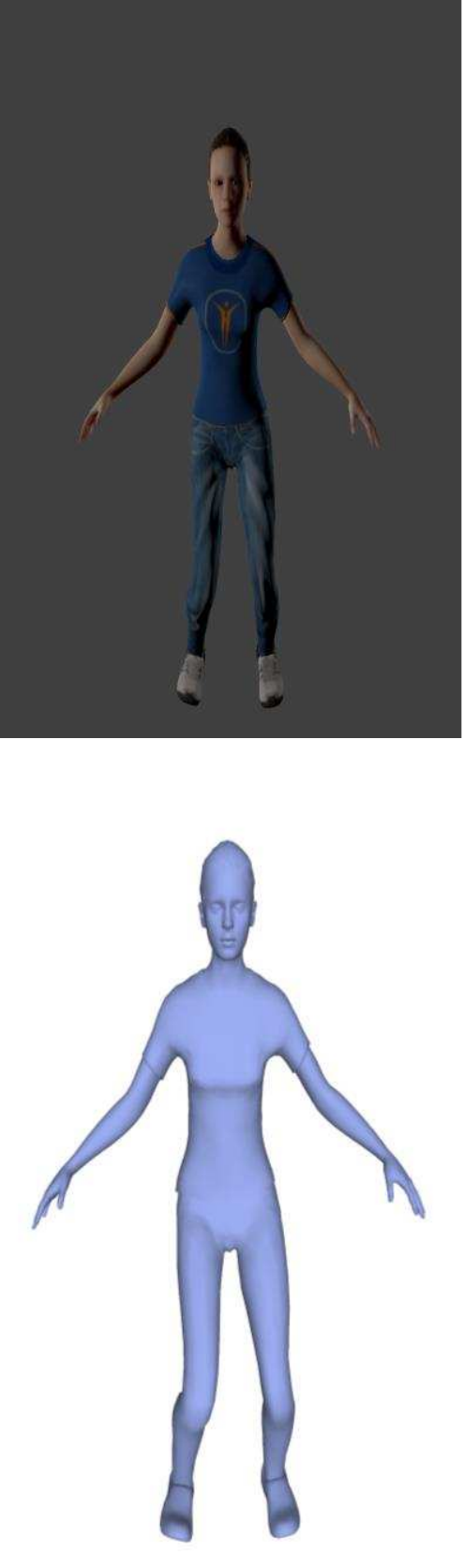}}{\#1}\hspace{0.5pt}
    \stackunder{\includegraphics[width= 0.16\linewidth,height=7cm]{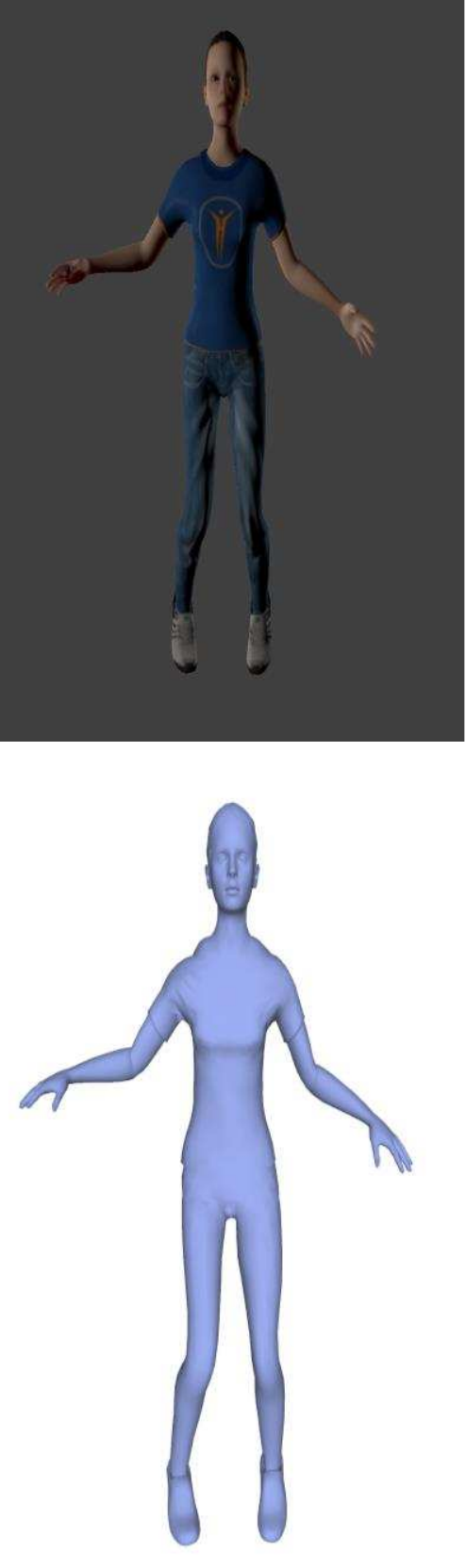}}{\#12}\hspace{0.5pt}
    \stackunder{\includegraphics[width= 0.16\linewidth,height=7cm]{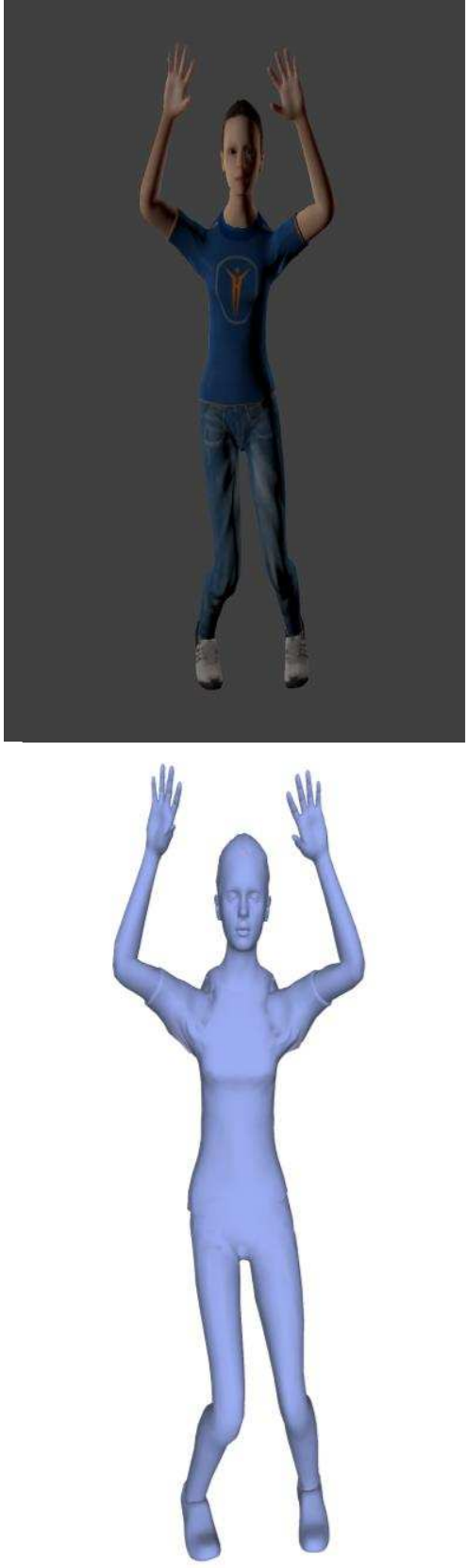}}{\#44}\hspace{0.5pt}
    \stackunder{\includegraphics[width= 0.16\linewidth,height=7cm]{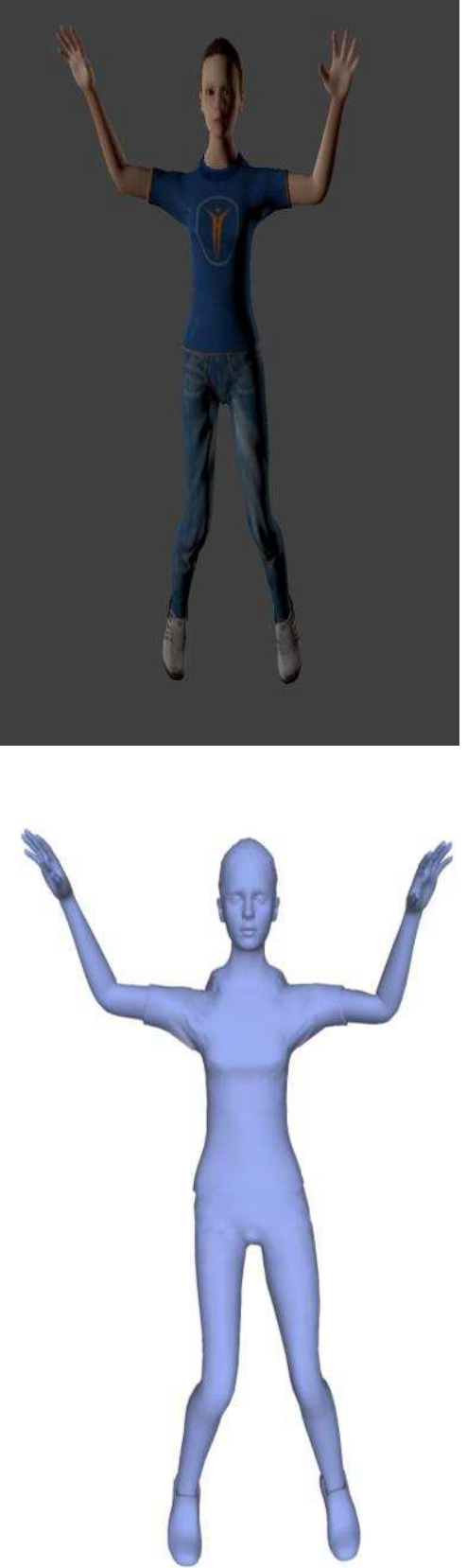}}{\#52}\hspace{0.5pt}
    \stackunder{\includegraphics[width= 0.16\linewidth,height=7cm]{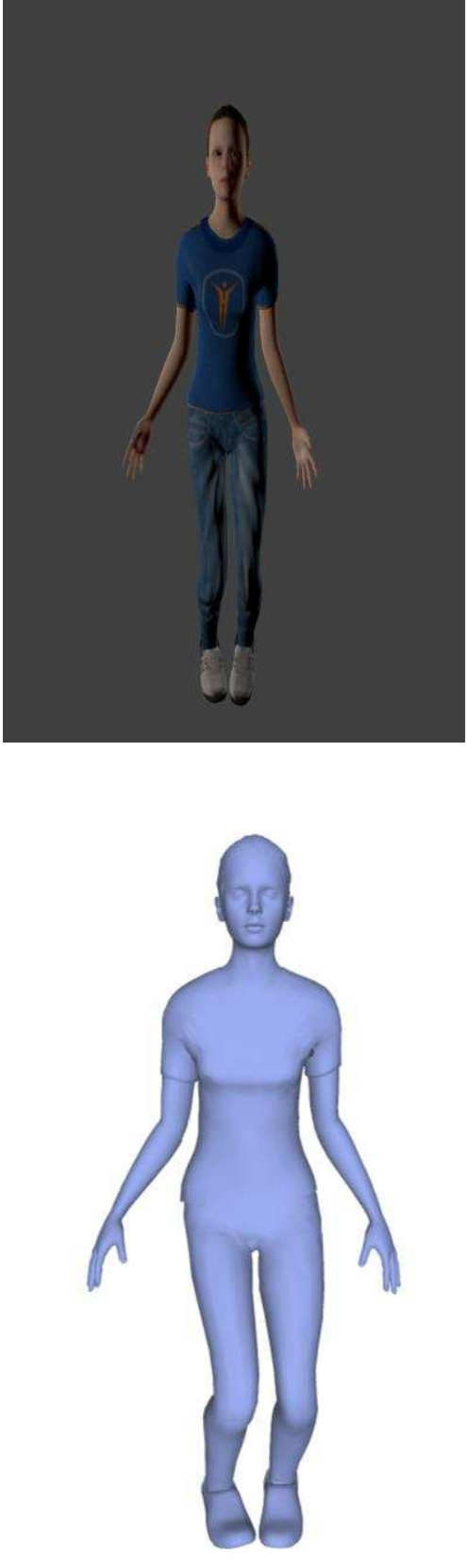}}{\#64}\hspace{0.5pt}
    \stackunder{\includegraphics[width= 0.16\linewidth,height=7cm]{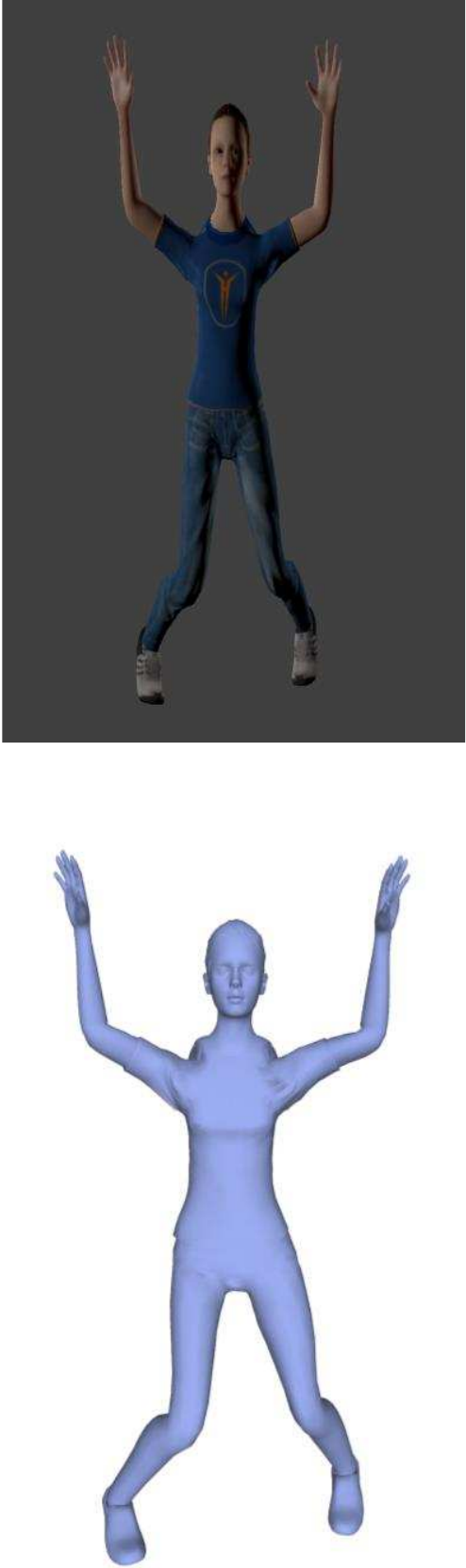}}{\#75}
    \end{center}
    \caption{Qualitative results of motion tracking from the \textquoteleft Exercise\textquoteright \hspace{1pt} data sequence from our dataset. The upper row shows images of different frames and the lower row shows the respective deformed 3D model. The frame index is shown below each image.}
   \label{QualitativeExercise}
\end{figure*}

\begin{figure*}[!htb]
    \begin{center}
    \stackunder{\includegraphics[width= 0.16\linewidth,height=7cm]{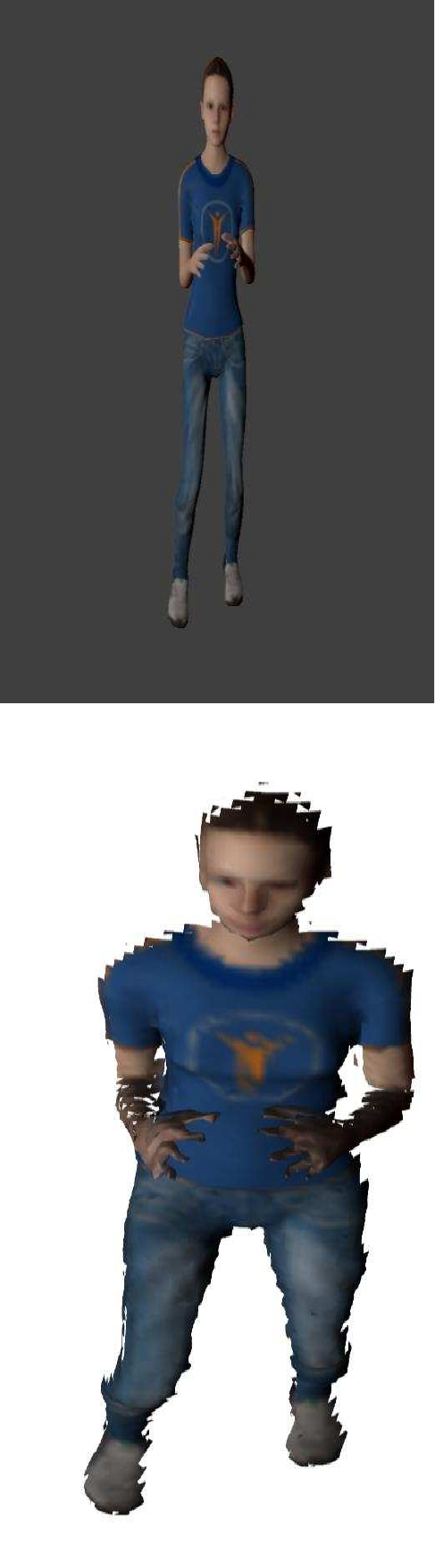}}{\#22}\hspace{0.5pt}
    \stackunder{\includegraphics[width= 0.16\linewidth,height=7cm]{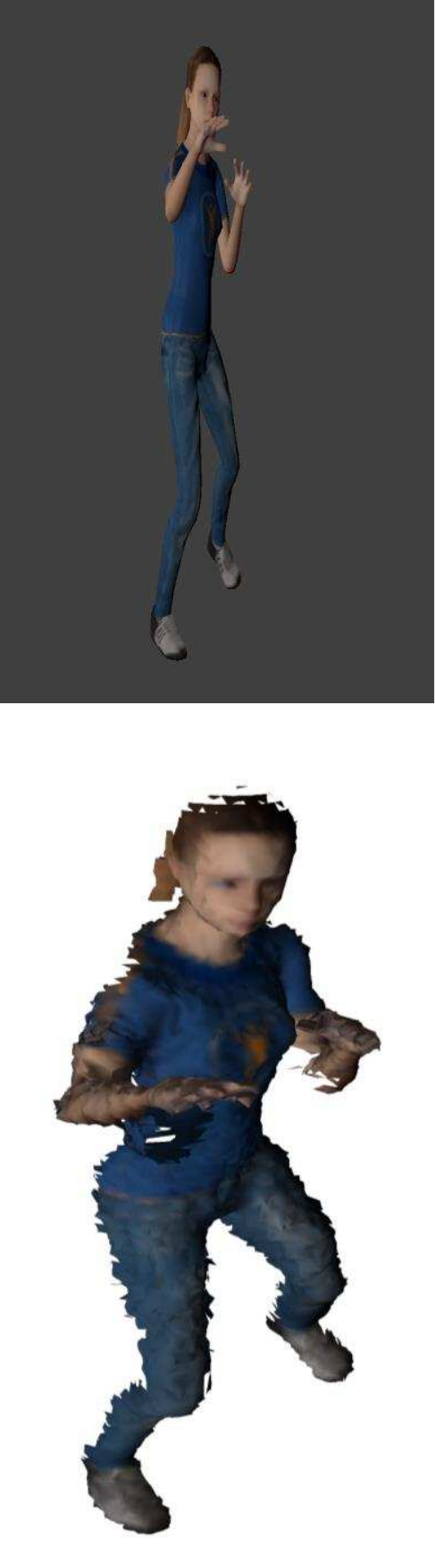}}{\#35}\hspace{0.5pt}
    \stackunder{\includegraphics[width= 0.16\linewidth,height=7cm]{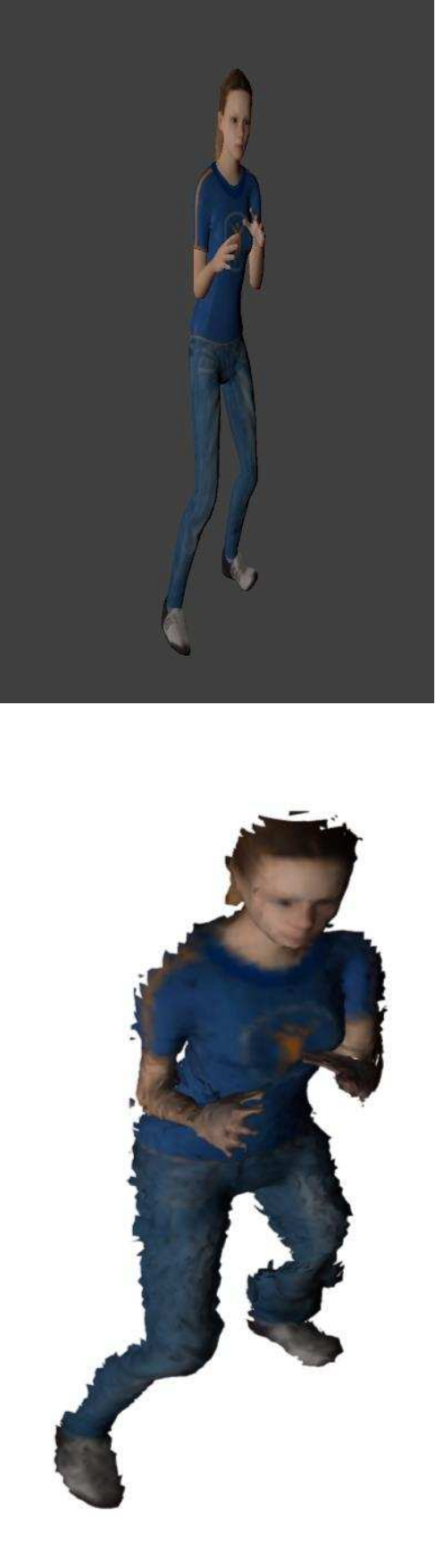}}{\#42}\hspace{0.5pt}
    \stackunder{\includegraphics[width= 0.16\linewidth,height=7cm]{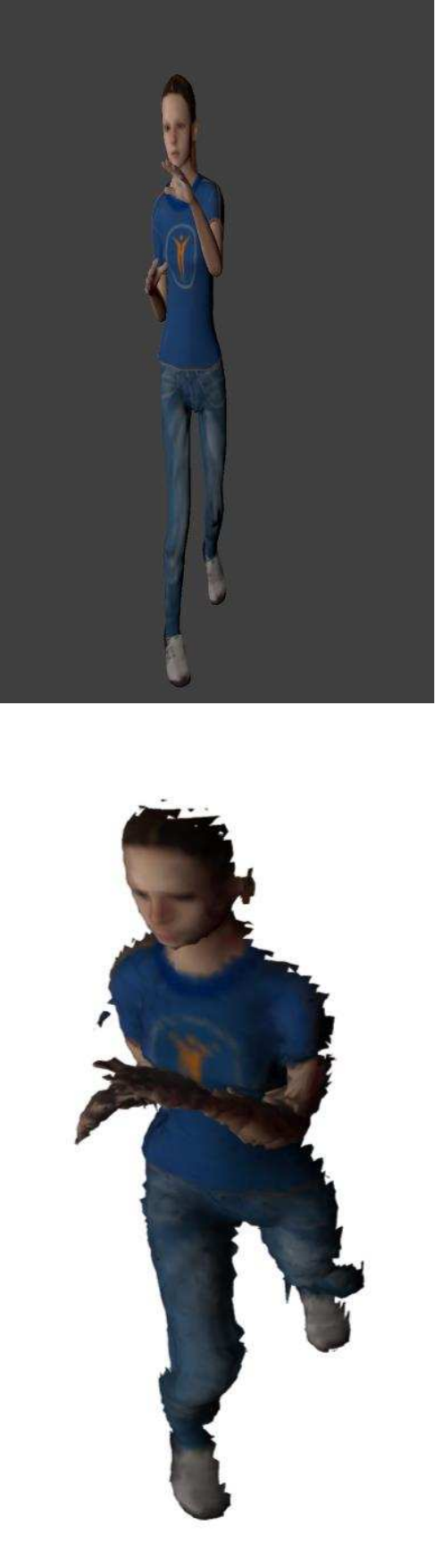}}{\#60}\hspace{0.5pt}
    \stackunder{\includegraphics[width= 0.16\linewidth,height=7cm]{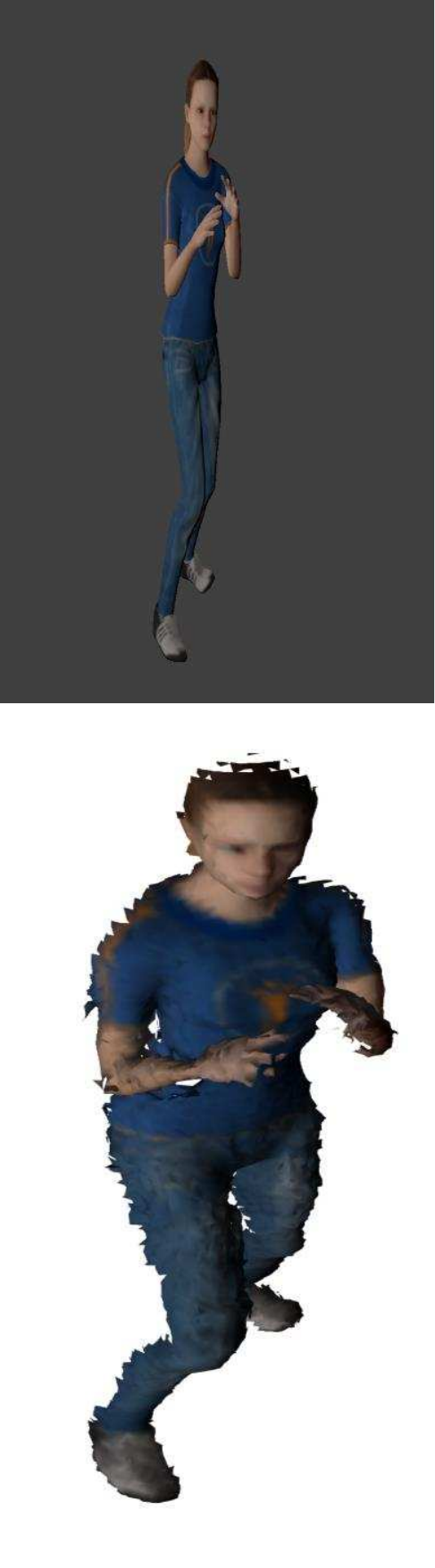}}{\#75}\hspace{0.5pt}
    \stackunder{\includegraphics[width= 0.16\linewidth,height=7cm]{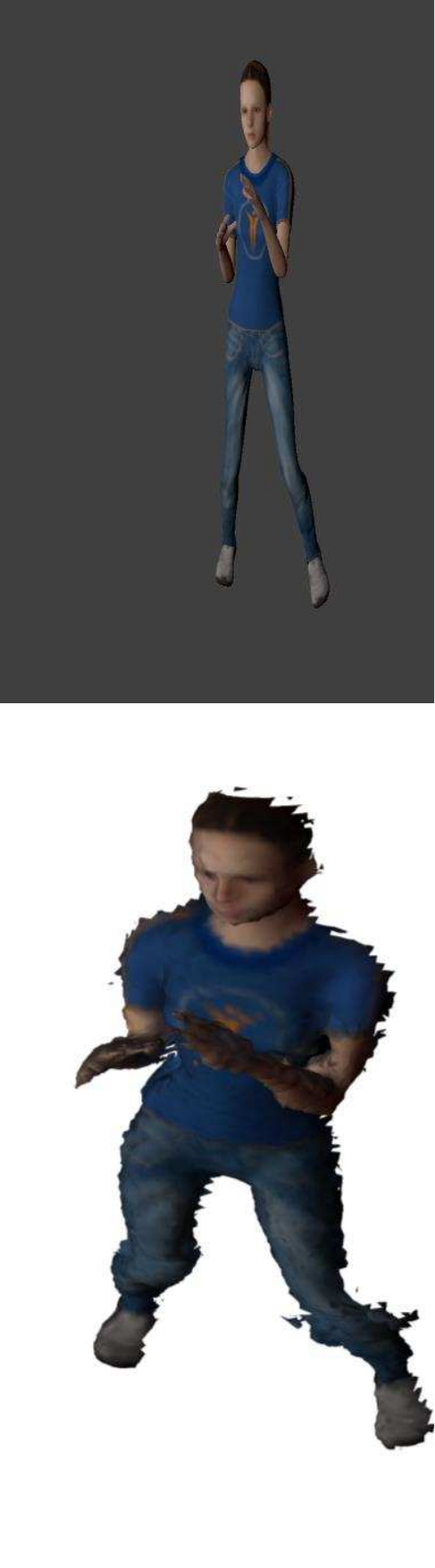}}{\#87}
    \end{center}
    \caption{Qualitative results of live 3D reconstruction from the \textquoteleft Boxing\textquoteright \hspace{1pt} sequence. The upper row shows images of different frames and the lower row shows the respective 3D reconstructions. The frame index is shown below each image.}
    \label{QualitativeBoxingRecon}
\end{figure*}

\subsection{Advantages of using the puppet model's rigid transformations for tracking initialization} \label{PuppetBasedInitialization}

At first, we evaluate the advantages of using the puppet model's rigid transformations for tracking initialization. In order to test this, we take $1^{st}$ and $7^{th}$ frames from \textquoteleft Exercise\textquoteright \hspace{1pt} and run non-rigid tracking in two cases. In the first case the tracking is started using the puppet model's rigid transformations and in the second case tracking is started without any initialization. Figure \ref{fig:QualitativeComparison} shows a qualitative comparison of 3D reconstruction results from both approaches. 
We can see that tracking failed in hand region of the second case and results in noise in the 3D reconstruction. Table \ref{tbl:trackingFromPuppetModel} gives a qualitative comparison between the both cases. We estimate MAE of point to plane distance from the ground truth geometry for each iteration. The second, third, and forth columns of Table \ref{tbl:trackingFromPuppetModel} show the iteration number, mean and standard deviation of errors from each case. We also show the Hausdroff distance \cite{aspert2002mesh} and the number of outliers in both cases. A point is selected to be an outlier if the point to mesh distance is more than 5mm. From the Table \ref{tbl:trackingFromPuppetModel} it can be observed that the case of tracking started using puppet model's rigid transformations have less error and less outliers in all instances. Note that state-of-the-art approaches \cite{yu122017bodyfusion, l0Norigid2015} always start tracking without any initialization.

\subsection{Comparison with state-of-the-art approaches}
We compared our approach with DynamicFusion \cite{newcombe2015dynamicfusion} and BodyFusion \cite{yu122017bodyfusion}. Figures \ref{Quant_Boxing} and \ref{Quant_PunchStrike} shows the qualitative comparison using the \textquoteleft boxing\textquoteright \hspace{1pt} and \textquoteleft Punch Strike\textquoteright \hspace{1pt} sequences respectively. Here the MAE error from each frame is plotted against the frame index. From Figures \ref{Quant_Boxing} and \ref{Quant_PunchStrike} we can observe that MAE error in DynamicFusion and BodyFusion goes beyond 0.1 mm within the first 50 frames whereas our approach always maintain an  MAE error below 0.02mm throughout the frame sequence. Figure \ref{Qual_Boxing} shows the qualitative comparison with state-of-the-art. From Figure \ref{Qual_Boxing} we can observe that both DynamicFusion and BodyFusion fail to reconstruct and errors accumulate, whereas our approach can reconstruct even with sudden articulated movements. 

\subsection{Qualitative results of non-rigid motion tracking}
Our approach uses a human model for motion tracking. The state-of-the-art approaches \cite{l0Norigid2015,liao2009modeling} use local optimization for tracking. Therefore the number of iterations usually depends upon the motion of the object, faster movements required more iterations. Moreover since these approaches use local optimization they may get stuck in local minima. Our approach uses skeleton joints for tracking. Therefore our approach can track any movements within four iterations irrespective of the movement speed. Qualitative results of motion tracking from \textquoteleft boxing\textquoteright\hspace{1pt} and \textquoteleft exercise\textquoteright\hspace{1pt} data sequences are shown in Figure \ref{QualitativeBoxing} and \ref{QualitativeExercise}.

\subsection{Qualitative results of non-rigid 3D reconstruction}
Similar to state-of-the-art approaches the proposed method 3D reconstructs a moving human subject over time. Because our approach can track sudden movements it can 3D reconstruct human subjects even with fast articulated movements. Qualitative reconstruction results from the \textquoteleft Boxing\textquoteright\hspace{1pt} and \textquoteleft Exercise\textquoteright\hspace{1pt} data sequences are shown in Figure \ref{QualitativeExerciseRecon} and  Figure \ref{QualitativeBoxingRecon}
respectively.

%% file: sections/conclusion.tex
\section{Conclusion}

We proposed a novel framework for non-rigid 3D reconstruction of human subjects that uses skeleton prior at each frame for tracking, with the help of a puppet model. This enables the proposed approach tracks sudden articulated movements without the need of any extra optimization iterations. The aligned puppet model provides correct correspondences for non-rigid reconstruction. We have demonstrated that our approach is more robust when faced with sudden articulated motions, and provides better reconstruction. We also contributed a synthetic dataset which provides ground truth for frame-by-frame geometry and skeleton joints of human subjects for evaluating non-rigid 3D reconstruction approaches of human subjects.